%% file: main.tex
\documentclass[11pt,a4paper,logo]{googledeepmind}

\microtypesetup{tracking=false}

\setleftlogo[180pt]{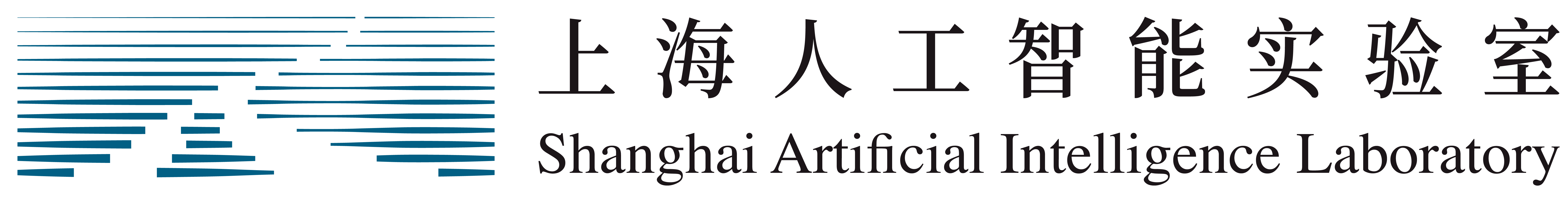}
\setrightlogo[180pt]{}

\renewcommand{\GDMRightLogo}{\mbox{}}

\usepackage[section]{placeins}
\usepackage{flafter}
\usepackage{subcaption}
\usepackage[
    natbib=true,
    backend=biber,
    style=numeric,
    sorting=none
]{biblatex}
\AtEveryBibitem{\clearfield{month}}
\AtEveryBibitem{\clearfield{day}}
\usepackage{csquotes}

\usepackage{pdflscape}
\usepackage{amsthm}

\usepackage{textcomp}
\usepackage{rotating}
\usepackage{setspace}
\usepackage{soul}
\usepackage{multicol}
\usepackage{caption}
\usepackage{array}
\usepackage{longtable}
\usepackage{multirow}
\usepackage{amsmath}
\usepackage{siunitx}
\usepackage{enumitem}
\usepackage{float}
\usepackage{algorithm}
\usepackage{algpseudocode}
\usepackage{seqsplit}
\usepackage{framed}
\usepackage{tikz}
\usepackage{listings}
\usepackage{colortbl}
\usepackage{wrapfig}
\usepackage[most]{tcolorbox}
\usepackage{pdfpages}
\tcbuselibrary{listingsutf8}
\usepackage{ragged2e}
\usepackage{makecell}
\usepackage{adjustbox}
\usepackage[bottom,symbol]{footmisc}
\usepackage{tocloft}
\usepackage{etoolbox}

\newcommand{\projectname}{Sage-Taiyi}

\title{\projectname: A Metacognitive Architecture for Scientific Discovery }
\author[1]{\projectname{} Team, Shanghai Artificial Intelligence Laboratory}

\definecolor{taiyiblue}{RGB}{70,123,157}
\definecolor{revblue}{RGB}{0,84,204}

\sethlcolor{green!14}

\newcolumntype{Y}{>{\RaggedRight\arraybackslash}X}

\makeatletter
\renewcommand{\maketitle}{%
    \thispagestyle{firststyle}%
    \begingroup
    \setlength{\parskip}{0pt}%
    \raggedright
    \begin{minipage}{\dimexpr\linewidth-23pt\relax}
    {\titlefont \@title\par}%
    \vspace{10pt}%
    {\@author\par}%
    \end{minipage}%
    \endgroup
    \par\vspace{6pt}%
    \abscontent
}
\makeatother

\hypersetup{
    colorlinks=true,
    linkcolor=blue,
    citecolor=blue,
    filecolor=black,
    urlcolor=blue
}
\definecolor{AbstractBgColor}{HTML}{F4F7FB}

\hypersetup{bookmarksdepth=3}

\cftsetpnumwidth{2em}
\cftsetrmarg{3em}
\defbibheading{bibintoc}[\refname]{%
    \section*{#1}%
    \phantomsection
    \addcontentsline{toc}{section}{#1}%
}

\begin{document}
\pagenumbering{roman}
\sloppy
\thispagestyle{firststyle}

\begin{tcolorbox}[
    colback=AbstractBgColor,
    colframe=AbstractBgColor,
    arc=5pt,
    auto outer arc,
    boxrule=0pt,
    left=8pt, right=8pt,
    top=4pt, bottom=4pt,
    parbox=false,
    width=\textwidth,
    before skip=-800pt,
    after skip=0pt,
    enlarge top by=-12pt
]
\begin{abstract}
We introduce \projectname{}, a metacognition-inspired architecture that uses accumulated scientific understanding to guide what to investigate next and how to investigate it. Through explicit monitoring and control, the system revises its understanding of evidence, assumptions, and unresolved questions, then uses these assessments to adapt research targets and select reusable reasoning methods. On Taiyi-Bench, an AI co-scientist ideation benchmark comprising 265 tasks from 50 papers across ten disciplines, \projectname{} ranks first on all eight metrics of historical rediscovery and open-ended ideation, achieving an Overall score of 54.0 compared with 47.9 for the strongest baseline, AI Scientist-v2. It also performs well on two external scientific ideation benchmarks. Five expert-guided collaborations further connect its proposals to experimental and computational validation across chemistry, materials science, and engineering. In chemistry, reaction-condition optimization raises the yield of cobalt-catalyzed carbonylation from trace amounts to 84\% and photocatalytic allylic amination from 13\% to 69\%. These results support metacognitive regulation as an approach to scientific ideation and its progression toward practical validation.
\end{abstract}

\maketitle
\par\smallskip
{\small\centering\hypersetup{urlcolor=taiyiblue}%
 \href{https://alpha-innovator.github.io/Sage-Taiyi/}{%
   \tikz[baseline=-0.5ex, line width=0.45pt]{
     \draw (0,0) circle[radius=0.65ex];
     \draw (0,0) ellipse[x radius=0.28ex, y radius=0.65ex];
     \draw (-0.65ex,0) -- (0.65ex,0);
   }\enspace\textbf{Homepage}}%
 \quad\textcolor{taiyiblue!45}{\textbar}\quad
 \href{https://discovery.intern-ai.org.cn/token-plan/home?tabIndex=1}{%
   \tikz[baseline=-0.5ex, line width=0.45pt]{
     \draw[line join=round] (0.15ex,0.7ex) -- (-0.45ex,-0.05ex)
       -- (0.05ex,-0.05ex) -- (-0.15ex,-0.7ex)
       -- (0.5ex,0.15ex) -- (0,0.15ex) -- cycle;
   }\enspace\textbf{Token Plan}}\footnotemark
 \par}
\end{tcolorbox}
\footnotetext{\textcolor{taiyiblue}{Sage-A1.5-Taiyi} brings Sage-Taiyi's Flash mode to Intern InkStone through the Token Plan. Built on Sage-A1.5, it advances the model's scientific discovery capabilities and accelerates the exploration of new research ideas.}

\par\vspace{34pt}
\begingroup
\hypersetup{linkcolor=black,linktoc=all}
\small
\setlength{\parskip}{0pt}
\setlength{\cftaftertoctitleskip}{8pt}
\setlength{\cftbeforesecskip}{4pt}
\setlength{\cftbeforesubsecskip}{1pt}
\setlength{\cftbeforesubsubsecskip}{0pt}
\pdfbookmark[1]{Contents}{report-contents}
\tableofcontents
\endgroup
\clearpage
\pagenumbering{arabic}

\input{section/introduction}
\input{section/method}
\input{section/experiments}
\input{section/related_work}
\input{section/conclusion}

\FloatBarrier
\printbibliography[heading=bibintoc]

\clearpage
\appendix
\input{section/appendix}

\end{document}

%% file: section/introduction.tex
\section{Introduction}
\label{sec:introduction}

As Carl Sagan observed, \textit{“Science is a way of thinking much more than it is a body of knowledge.”} Scientific discovery involves more than producing new hypotheses or accumulating findings. It unfolds through an iterative loop: scientific search generates and tests ideas, scientific understanding develops through the interpretation of the resulting findings, and this understanding guides subsequent search~\cite{klahrdunbar1988dual}. Researchers move within and between hypothesis and experiment spaces around a scientific question, proposing explanations, designing tests, and interpreting experimental, computational, or observational feedback~\cite{klahr2000exploring}. Interpreting this feedback develops a working understanding of the problem: which judgments are supported, which conflicts remain unresolved, under which conditions findings apply, and which regularities may transfer across cases. As understanding changes, researchers revise hypotheses, reformulate questions, redesign experiments, and reassess which direction merits further effort. Classical problem-solving research describes how representations, operators, and heuristics organize this search and guide subsequent action~\cite{newellsimon1972human,langley1987scientific}. The central design question is how to retain the understanding developed during search and use it to select and adapt representations, operators, and heuristics, so that search experience informs subsequent research decisions.

\begin{figure}[!t]
    \centering
    \includegraphics[width=\linewidth]{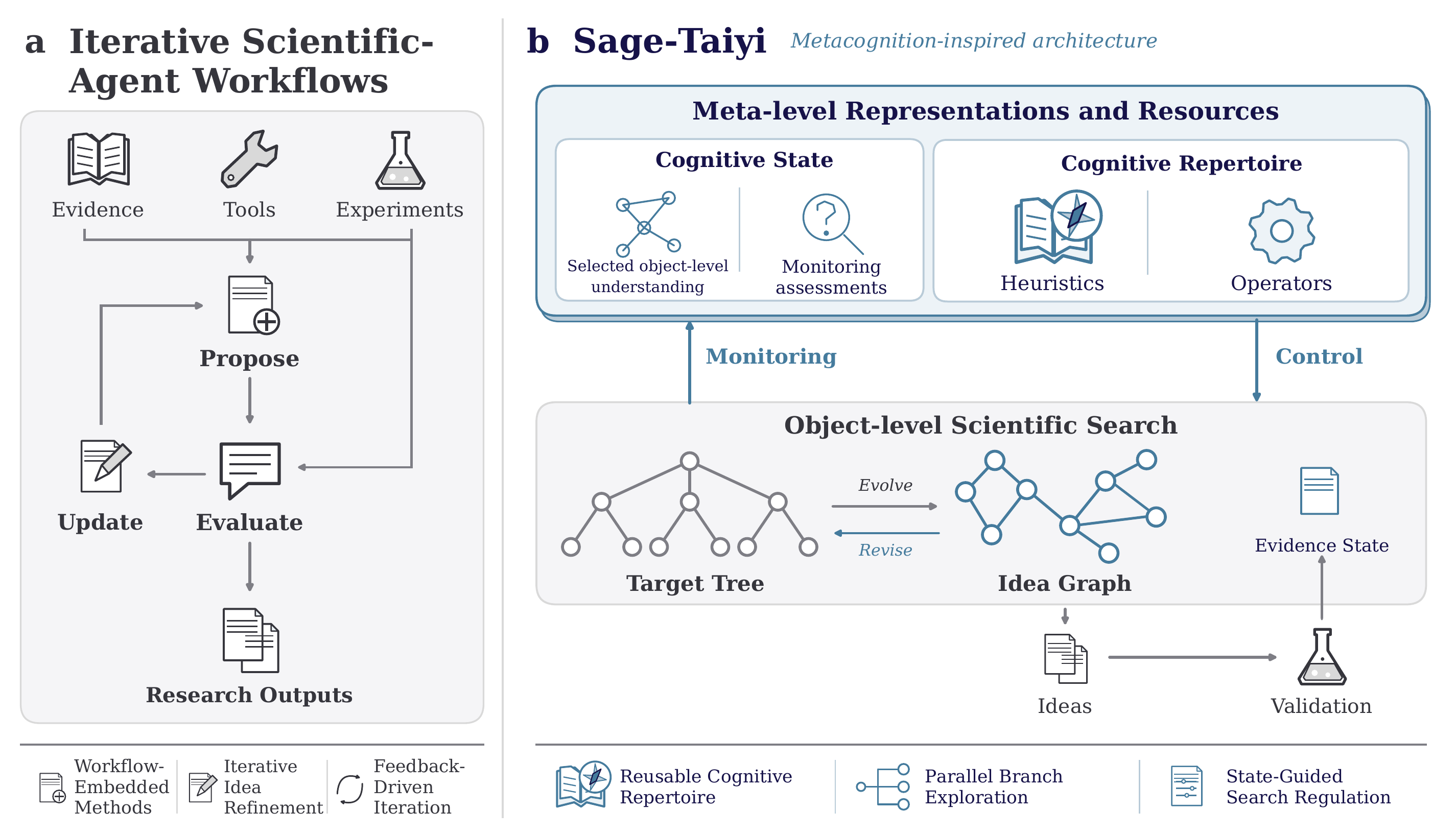}
    \caption{\textbf{(a)} Iterative scientific-agent workflows organize discovery through cycles of proposing, evaluating, and updating ideas, using evidence, tools, and experiments to guide refinement and produce research outputs. \textbf{(b)} \projectname{} connects object-level scientific search with meta-level cognitive regulation through monitoring and control. At the object level, a Target Tree organizes search targets, while an Idea Graph captures ideas as they are generated and refined. At the meta level, a Cognitive Repertoire supplies methodological heuristics and Cognitive Operators, while a Cognitive State maintains a selective representation of object-level scientific understanding alongside monitoring-derived assessments that guide subsequent search. Monitoring updates the Cognitive State from search outcomes; control uses the updated state to select targets and methods that guide subsequent scientific search. Selected ideas can proceed to validation, whose results return to the Evidence State. The two panels highlight different architectural emphases rather than mutually exclusive capabilities.}
    \label{fig:intro-compare}
\end{figure}

Recent AI systems have extended one-shot idea generation into iterative scientific workflows. The AI Scientist connects ideation with experiments, analysis, and paper writing~\cite{lu2026aiscientist}; Co-Scientist organizes hypothesis generation, critique, and improvement~\cite{gottweis2026coscientist}; and Robin combines multi-agent reasoning with experimental feedback~\cite{ghareeb2026robin}. Iteration alone, however, does not ensure that findings from different branches are retained with their evidence and applicability conditions, or that these findings affect later method selection. When iteration depends mainly on feedback from the current round, earlier failures, unresolved conflicts, and shared dependencies can be overlooked, leading to repeated errors or further search along unproductive paths. Search can also become overly responsive to recent outcomes, delaying recognition of key uncertainties or a change in targets, representations, or reasoning methods. A scientific agent therefore needs to preserve understanding across the search and use it to regulate what to investigate next and how to investigate it.

This need motivates drawing on metacognition, a concept rooted in the study of human cognition. Often described as “thinking about thinking,” metacognition involves monitoring, and regulation of one's own cognitive processes~\cite{flavell1979metacognition}. Nelson and Narens distinguish an object level that performs a task from a meta level that maintains a dynamic model of the object-level cognition and regulates its process through monitoring and control~\cite{nelsonnarens1990metamemory}. Prior work examines how artificial systems can monitor and adjust their own problem-solving processes~\cite{cox2005metacognition}, complementary research on  strategy selection motivates matching reasoning methods to the task and current state~\cite{liedergriffiths2017strategy}. Viewed through the lens of metacognition, object-level search develops scientific understanding, while the meta level selectively represents this understanding and assesses its implications for further search, guiding decisions about which premises to examine, representations to revise, methods to apply, or branches to stop. We adopt this functional distinction as architectural inspiration for explicit search regulation.

We introduce \projectname{}, a metacognition-inspired architecture that uses accumulated scientific understanding to guide what to investigate next and how to investigate it. Through explicit monitoring and control, the system revises its understanding of evidence, assumptions, and unresolved questions, then uses these assessments to adapt research targets and select reusable reasoning methods. \projectname{} achieves leading performance on Taiyi-Bench, AgentIdeaBench, and IG-Arena. On Taiyi-Bench, an AI co-scientist ideation benchmark comprising 265 tasks from 50 papers across ten disciplines, it ranks first on all eight metrics of historical rediscovery and open-ended ideation, achieving an Overall score of 54.0 compared with 47.9 for the strongest baseline, AI Scientist-v2. Five expert-guided collaborations further connect its proposals to experimental and computational validation across chemistry, materials science, and engineering. In chemistry, reaction-condition optimization raises the yield of cobalt-catalyzed carbonylation from trace amounts to 84\% and photocatalytic allylic amination from 13\% to 69\%. These results support metacognitive regulation as an approach to scientific ideation and its progression toward practical validation.

%% file: section/method.tex
\definecolor{methodlabelfill}{RGB}{237,243,247}
\newcommand{\methodterm}[1]{%
    \begingroup
    \setlength{\fboxsep}{1pt}%
    \setlength{\fboxrule}{0.3pt}%
    \fcolorbox{taiyiblue!25}{methodlabelfill}{\strut\textbf{#1}}%
    \endgroup}

\section{Method}
\label{sec:method}

\projectname{} is a metacognition-inspired architecture for scientific discovery in which the search for ideas and the understanding of the scientific problem develop together. Section~\ref{sec:method-overview} defines the object and meta levels and how they interact. At the object level, scientific search constructs and transforms ideas across target branches (Section~\ref{sec:object-search}). At the meta level, a Cognitive State selectively represents object-level scientific understanding and records monitoring assessments, while a Cognitive Repertoire retains reusable methods (Section~\ref{sec:meta-representation}). Monitoring and control connect the two levels (Section~\ref{sec:monitoring-control}), and a research harness links ideas to planning, tool use, and external validation (Section~\ref{sec:real-world-feedback-method}).

\subsection{System Overview}
\label{sec:method-overview}

\projectname{} organizes scientific discovery into two levels, drawing on accounts of discovery as problem solving~\cite{newellsimon1972human,klahrsimon1999discovery} and of metacognition as monitoring and control of thinking~\cite{ackermanthompson2017metareasoning,fleming2024metacognition}.

\begin{figure}[htbp]
    \centering
    \includegraphics[width=\linewidth]{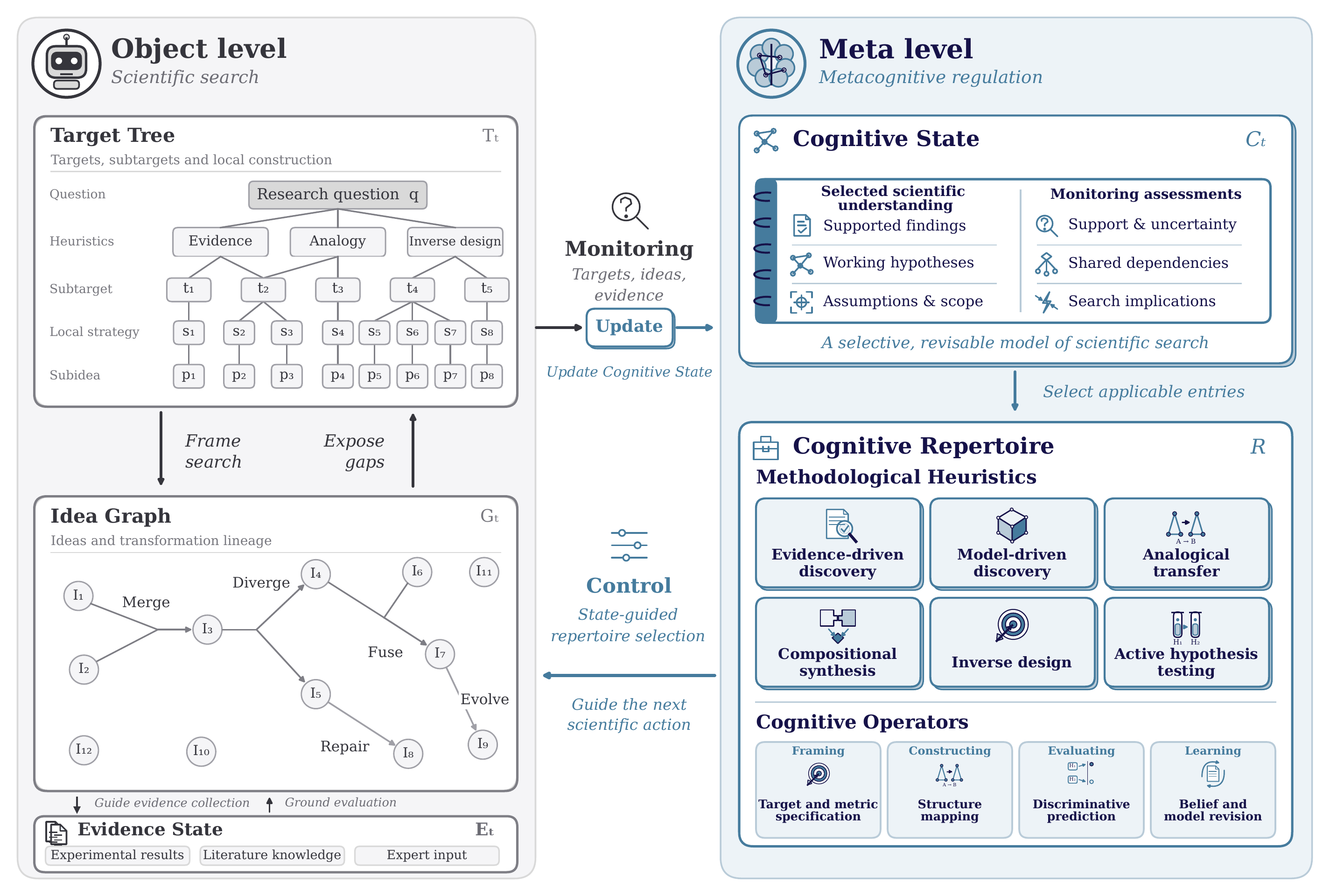}
    \caption{\textbf{\projectname{}'s two-level scientific search and monitoring--control loop.} The architecture comprises object-level scientific search and meta-level metacognitive regulation. The object level includes a Target Tree that organizes targets, subtargets, local strategies, and subideas; an Idea Graph that records ideas and their lineage; and an Evidence State that retains experimental results, literature knowledge, and expert input. The meta level includes a Cognitive State that maintains a selective representation of object-level scientific understanding alongside monitoring-derived assessments, and a Cognitive Repertoire that provides reusable methodological heuristics and Cognitive Operators. Monitoring and control coordinate the two levels: the Target Tree guides idea exploration, the Idea Graph reveals gaps encountered during exploration, and the Evidence State grounds evaluation. Monitoring uses these inputs to update the Cognitive State, assessing evidential support, uncertainty, shared dependencies, and their implications for search. Control uses the Cognitive State to select applicable heuristics and operators and guide subsequent scientific actions, enabling scientific understanding and the search for novel ideas to evolve together.}
    \label{fig:method-overview}
\end{figure}

The object level addresses the scientific problem itself, carrying out scientific search for a research question $q$. As in problem-space accounts of discovery, this search moves between knowledge states through local transformations~\cite{newellsimon1972human,klahrdunbar1988dual}. Its state is $X_t=(T_t,G_t,E_t)$: the Target Tree $T_t$ organizes targets and subtargets, the Idea Graph $G_t$ records ideas and their lineage, and the Evidence State $E_t$ retains experimental and computational results, literature knowledge, and expert input with their sources. Decomposing targets, generating, transforming, and evaluating ideas, and judging what the evidence supports are object-level operations, as is the integration of these judgments into scientific understanding.

The meta level assesses how scientific search is proceeding and decides how it should be adjusted~\cite{flavell1979metacognition}. It maintains two resources. The task-specific Cognitive State $C_t$ records what the search has established about the problem and what this implies for further search (Section~\ref{sec:cognitive-state}). The reusable Cognitive Repertoire $\mathcal{R}$ contains methodological heuristics, which organize research paths, and Cognitive Operators, which perform local reasoning transformations between knowledge states (Sections~\ref{sec:methodological-heuristics}--\ref{sec:cognitive-operators}). Selecting methods to regulate scientific search is a meta-level decision; applying them to develop targets or ideas is object-level search.

Monitoring and control connect the two levels~\cite{nelsonnarens1990metamemory}. Monitoring interprets search outcomes and new evidence to update $C_t$ and assess their implications for further search. Control uses $C_t$ to decide which target or idea to address, whether to continue, revise, redirect, or stop a branch, and which entries $\mathcal{R}_t\subseteq\mathcal{R}$ to use. Applying these decisions and selected methods updates the object-level search state:

\begin{equation}
q\longrightarrow
\left[
\underbrace{X_t=(T_t,G_t,E_t)}_{\text{Object-level search}}
\underset{\text{Control using }\mathcal{R}_t}{\overset{\text{Monitoring}}{\rightleftarrows}}
\underbrace{(C_t,\mathcal{R})}_{\text{Meta-level resources}}
\right]
\xrightarrow{\text{Review \& Execution}}\underbrace{o}_{\text{Output}}.
\label{eq:metacognitive-transition}
\end{equation}

The index $t$ marks logical state changes rather than fixed-duration rounds. A task ends with reviewed ideas or executable plans, or continues to execution and validation, whose results return to the Evidence State (Figure~\ref{fig:method-overview}).

% Keep the navigation anchor with this heading and its first subheading.
\Needspace{6\baselineskip}
\subsection{Object-Level Scientific Search}
\label{sec:object-search}

\subsubsection{Target Tree: Decomposing the Research Question}
\label{sec:target-decomposition}

The Target Tree decomposes the research question into more specific targets and subtargets. Its root is $q$, and each node is a target or subtarget that records its scientific requirements and completion criteria. Under a methodological heuristic selected by control, a target is decomposed into subtargets; each subtarget is paired with local strategies for addressing it, and each strategy yields subideas, partial solutions that can later be assembled into complete ideas. This structure follows problem-space accounts in which a problem is reduced to subproblems that are solved by applying operators~\cite{newellsimon1972human,klahrsimon1999discovery}. The tree is not fixed at the outset: as ideas develop, subtargets are added or revised while the root question and its constraints are retained (Section~\ref{sec:idea-evolution}).

\subsubsection{Idea Graph: Developing Ideas across Branches}
\label{sec:idea-evolution}

Ideas are the units that \projectname{} evaluates, transforms, and outputs. An idea is a complete proposal specifying a mechanism or design, supporting references, constraints, testable consequences, and the uncertainties to be examined. The Idea Graph records ideas as nodes and their transformations as edges. Ideas enter the graph by assembling subideas from the Target Tree or by direct generation from the research question, with or without a selected heuristic. Because the graph spans target branches, ideas developed for different targets can be compared and recombined.

Five transformations, informed by population-based search and conceptual combination~\cite{holland1992genetic,novikov2025alphaevolve,finkewardsmith1992creative,foster2015tradition}, produce new ideas. \methodterm{Merge} consolidates ideas that take similar approaches; \methodterm{Fuse} combines complementary mechanisms from different ideas after checking that their assumptions and interfaces are compatible; \methodterm{Diverge} keeps an idea and opens a substantively different branch from it; \methodterm{Repair} fixes a defective dependency or a missing interface; and \methodterm{Evolve} improves an idea while retaining its core mechanism. Over successive iterations, these transformations can refine ideas and combine strengths developed for different targets. As ideas develop, they can expose requirements that no current target covers, motivating new subtargets in the Target Tree. To keep this development traceable, each new idea is linked to its parents and to the evaluation that motivated it.

\subsubsection{Evidence State: Grounding Scientific Understanding}
\label{sec:evidence-understanding}

The Evidence State grounds scientific search in external information. It records experimental and computational results, literature knowledge, and expert input, linking each to its source, relevant ideas, and conditions. Using this evidence, the object level forms scientific judgments, for example which mechanisms are supported, under which conditions findings hold, and which explanations conflict. Each judgment retains its evidential grounds, allowing new evidence to inform revisions to relevant judgments and their dependencies while preserving unaffected alternatives. Judgments formed in one branch can also inform reasoning in others. Integrated across branches, these evidence-grounded judgments and their relations form the system's current scientific understanding of the problem~\cite{gickholyoak1983schema,thagard1989coherence}.

\subsection{Meta-Level Representations and Resources}
\label{sec:meta-representation}
\label{sec:cognitive-repertoire}

While the object level develops targets and ideas, the meta level keeps track of how this search is proceeding and adjusts its course. The meta level regulates scientific search with two resources. The Cognitive State represents the current search and is revised by monitoring as new findings or assessments warrant an update. The Cognitive Repertoire retains reusable method knowledge: methodological heuristics, which organize research paths, and Cognitive Operators, which transform knowledge states.

\subsubsection{Cognitive State: Representing the Current Search}
\label{sec:cognitive-state}

Following Nelson and Narens' account of the meta level as maintaining a dynamic model of object-level cognition~\cite{nelsonnarens1990metamemory}, the Cognitive State models the current search. It contains two kinds of content. The first is a selective, source-linked representation of object-level scientific understanding, retaining key findings, competing explanations, necessary assumptions, applicability conditions, and unresolved questions. The second consists of monitoring-derived assessments of the evidential status of key judgments, unresolved conflicts and uncertainties, and dependencies shared across ideas. The former selectively represents what the system currently understands about the scientific problem; the latter assesses how this understanding should guide further search. For example, when several ideas propose different mechanisms but depend on the same untested premise, monitoring records that they share a potential failure condition and that their agreement does not provide independent support.

The two kinds of content describe functions of the state rather than separate storage modules. Monitoring revises both as new evidence and evaluations of ideas emerge (Section~\ref{sec:monitoring}), preserving sources and keeping working hypotheses distinct from external evidence. Control uses the revised state to select methods from the repertoire and determine subsequent research actions; their application updates targets and ideas (Section~\ref{sec:control}).

\subsubsection{Methodological Heuristics: Organizing Research Paths}
\label{sec:methodological-heuristics}

\begin{figure}[H]
    \centering
    \includegraphics[width=\linewidth]{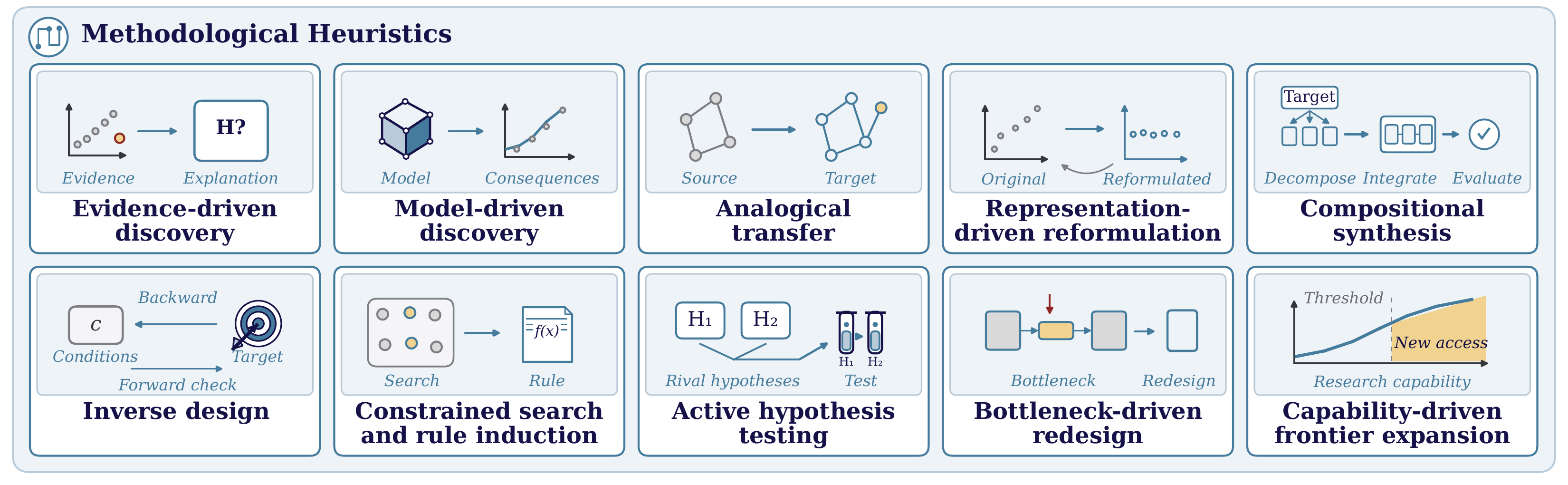}
    \caption{\textbf{Methodological heuristics.} The ten heuristics organize recurring strategies for scientific research grounded in evidence, models, analogies, representations, goals, and practical constraints.}
    \label{fig:methodological-heuristics}
\end{figure}

Methodological heuristics organize complete research workflows. Since the space of possible research paths is too large to search exhaustively, scientists in different disciplines develop characteristic research paradigms that lead them to workable solutions more efficiently~\cite{newellsimon1972human,langley1987scientific,kulkarni1988processes,darden1991theory}. Drawing on these accounts, \projectname{} organizes related reasoning principles into ten methodological heuristics for reusable research workflows (Figure~\ref{fig:methodological-heuristics} and Table~\ref{tab:methodologies}). Each heuristic specifies when it applies, how it decomposes a target into subtargets and local strategies, coordinates cognitive operators, and evaluates the resulting ideas.

These general heuristics are further instantiated with discovery methods summarized from the literature of specific research domains (Section~\ref{sec:repertoire-learning}). Control selects heuristics according to the research question and the current Cognitive State~\cite{liedergriffiths2017strategy}, and the selected heuristics guide the construction of the Target Tree (Section~\ref{sec:target-decomposition}).

\begin{table}[H]
    \centering
    \footnotesize
    \renewcommand{\arraystretch}{1.10}
    \setlength{\fboxsep}{2pt}
    \setlength{\fboxrule}{0.3pt}
    \newcommand{\heuristiclabel}[1]{%
        \fcolorbox{taiyiblue!25}{methodlabelfill}{%
            \begin{varwidth}[t]{\dimexpr\linewidth-2\fboxsep-2\fboxrule\relax}%
                \raggedright #1%
            \end{varwidth}}}
    \caption{Methodological heuristics and their roles in scientific search.}
    \label{tab:methodologies}
    \begin{tabularx}{\linewidth}{@{}>{\RaggedRight\arraybackslash}p{0.30\linewidth}Y@{}}
        \toprule
        Methodological heuristic & Methodological role \\
        \midrule
        \heuristiclabel{Evidence-driven discovery} & Construct explanations from observations, anomalies, or failed predictions, and identify subsequent tests~\cite{klahrdunbar1988dual,langley1987scientific}. \\[2pt]
        \heuristiclabel{Model-driven discovery} & Vary assumptions or conditions in a theory or model, derive new consequences, and examine their scope and testable predictions~\cite{nersessian1999modelbased}. \\[2pt]
        \heuristiclabel{Analogical transfer} & Align the relational structures of source and target problems, and transfer mechanisms under target-domain constraints~\cite{gentner1983structuremapping,holyoakthagard1995mentalleaps}. \\[2pt]
        \heuristiclabel{Representation-driven reformulation} & Reformulate entities, variables, or relations to open new solution paths, then map results back to the original problem~\cite{amarel1981representations,ohlsson1992insight,knoblich1999constraint}. \\[2pt]
        \heuristiclabel{Compositional synthesis} & Decompose a target into functions, select and integrate compatible components, and evaluate the resulting system~\cite{bechtel2010discovering,finkewardsmith1992creative}. \\[2pt]
        \heuristiclabel{Inverse design} & Infer conditions and subtargets from a required outcome, construct a design, and validate it through forward evaluation~\cite{newellsimon1972human}. \\[2pt]
        \heuristiclabel{Constrained search and rule induction} & Explore ideas in a feasible space, and induce testable regularities from successful and unsuccessful instances~\cite{langley1987scientific,klahrdunbar1988dual}. \\[2pt]
        \heuristiclabel{Active hypothesis testing} & Select observations or interventions that distinguish competing explanations, and revise hypotheses using the results~\cite{oaksfordchater1994selection,steyvers2003causal}. \\[2pt]
        \heuristiclabel{Bottleneck-driven redesign} & Identify the active constraint on overall system performance, redesign around it, and examine whether the bottleneck shifts~\cite{rosenberg1969focusing}. \\[2pt]
        \heuristiclabel{Capability-driven frontier expansion} & Use advances in measurement, experimentation, or computation to investigate previously inaccessible questions~\cite{price1984instrumentalities,nersessian2006distributed}. \\[2pt]
        \bottomrule
    \end{tabularx}
\end{table}

\subsubsection{Cognitive Operators: Transforming Knowledge States}
\label{sec:cognitive-operators}

\begin{figure}[H]
    \centering
    \includegraphics[width=\linewidth]{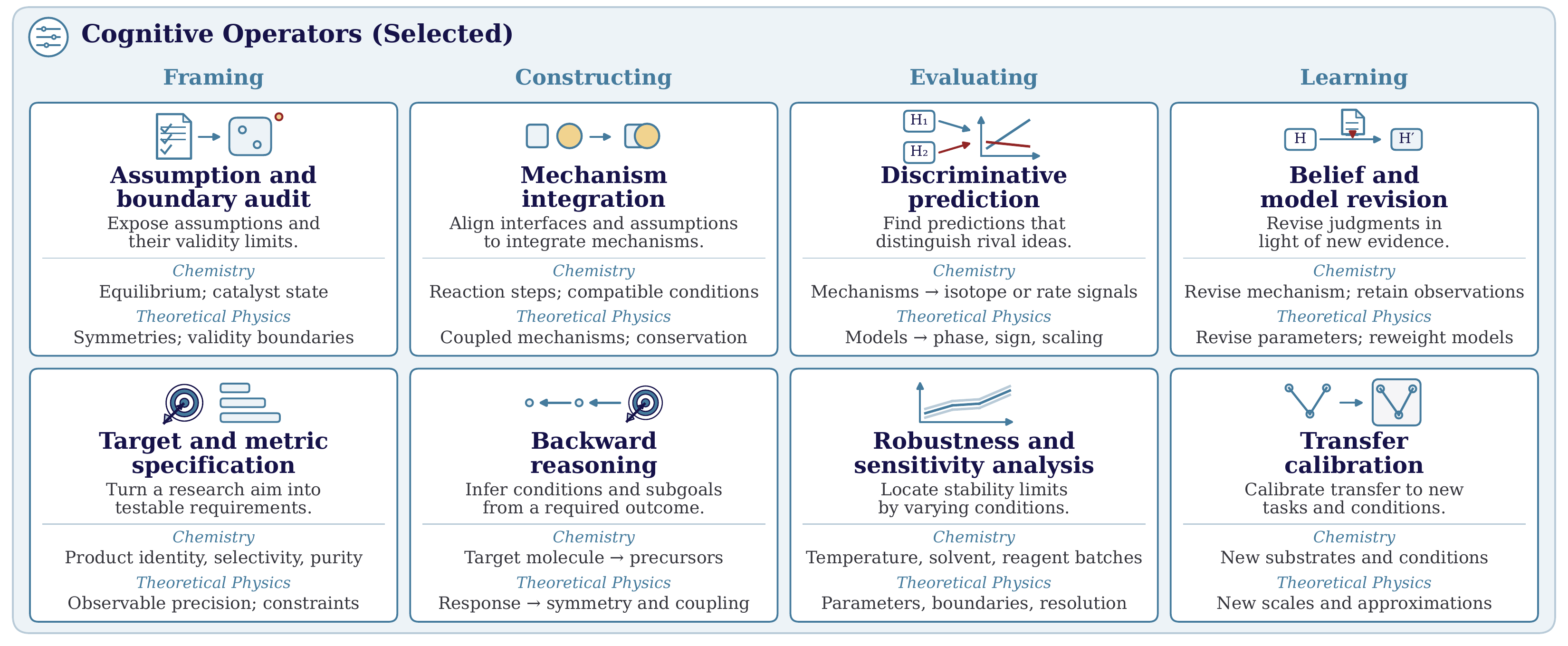}
    \caption{\textbf{Selected Cognitive Operators.} Eight representative operators illustrate local reasoning transformations across Framing, Constructing, Evaluating, and Learning. Each card states the operator's function and gives examples of its specialization in chemistry and theoretical physics.}
    \label{fig:cognitive-operators}
\end{figure}

Drawing on problem-space accounts of human problem solving~\cite{newellsimon1972human}, \projectname{} defines Cognitive Operators as local reasoning transformations reusable across methodological workflows. Backward reasoning infers conditions or subgoals from an outcome; inverse design also constructs and validates a design. Mechanism integration aligns existing mechanisms' interfaces and assumptions to form a candidate; compositional synthesis also decomposes the target and evaluates the system. Operators act on targets in the Target Tree, ideas and evaluations in the Idea Graph, representations and assessments in the Cognitive State, and methods in the Cognitive Repertoire. Informed by process accounts of creative thinking~\cite{mumfordmcintosh2017processes,benedek2023memory}, we organize operators by their principal function: \methodterm{Framing} specifies targets and their requirements; \methodterm{Constructing} develops and transforms ideas; \methodterm{Evaluating} examines their evidential support and testable consequences; and \methodterm{Learning} uses search outcomes to revise the Cognitive State and refine reusable methodological knowledge in the Cognitive Repertoire. These categories allow overlap, and an operator may act on multiple structures. Each operator specifies applicability conditions, a reasoning procedure, and expected outputs; control selects applicable operators using the task and Cognitive State. Figure~\ref{fig:cognitive-operators} illustrates selected operators and their specialization in chemistry and theoretical physics.

\subsubsection{Domain Instantiation and Feedback Learning}
\label{sec:repertoire-learning}

\begin{figure}[!htb]
    \centering
    \includegraphics[width=0.8\linewidth]{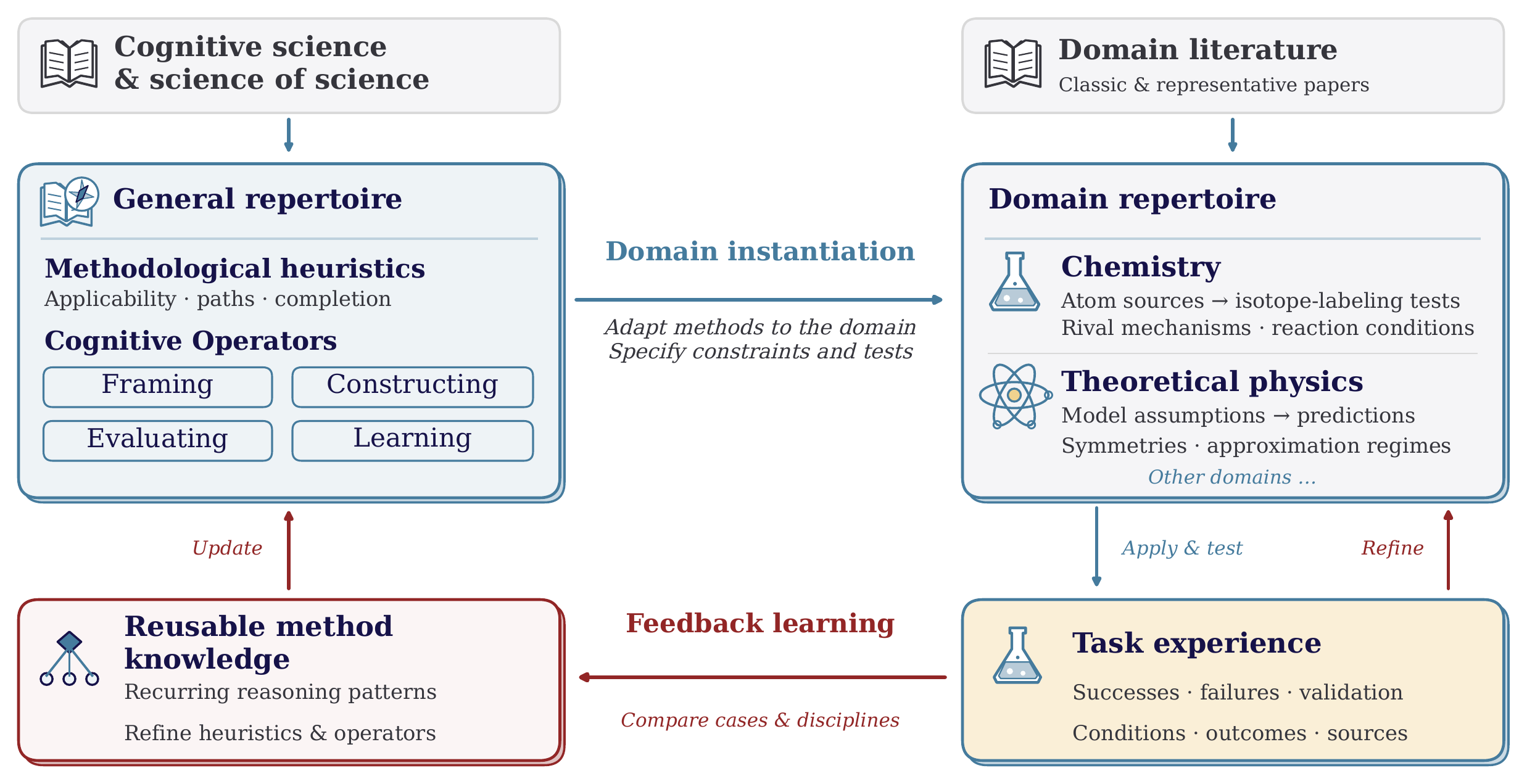}
    \caption{\textbf{Domain instantiation and feedback learning.} General heuristics and operators are specialized using domain literature. Task feedback refines domain entries; comparisons across cases and disciplines yield reusable knowledge that revises the general repertoire. Transferred experience retains its sources and scope and is re-examined in each new domain.}
    \label{fig:repertoire-learning}
\end{figure}

The Cognitive Repertoire combines general heuristics and operators inspired by cognitive science~\cite{newellsimon1972human} and the science of science~\cite{foster2015tradition} with domain-specific research methods. Figure~\ref{fig:repertoire-learning} summarizes their development through domain instantiation and feedback learning.

\textbf{Domain instantiation} analyzes classic and representative papers to map general reasoning procedures to domain objects, mechanisms, constraints, and validation requirements. This mapping preserves relational structures while grounding their application in disciplinary knowledge and experimental conditions~\cite{gentner1983structuremapping,nersessian2006distributed}. For example, active hypothesis testing becomes a concrete chemistry procedure when two mechanisms predict different sources for a product atom: isotope-labeling one reactant and examining the product provides a test of their predictions. Entries retain their sources and applicability conditions, distinguishing reported procedures from analytical reconstruction.

\textbf{Feedback learning} uses successes, failures, and validation outcomes to refine these entries, drawing inspiration from case-based learning through reuse, testing, revision, and retention~\cite{aamodtplaza1994casebased}. Comparisons across cases and disciplines identify recurring reasoning patterns that inform revisions to general heuristics and operators, following the principle of schema induction from concrete examples~\cite{gickholyoak1983schema}. Revised methods can then be instantiated in further domains, with their operating conditions re-examined during transfer, closing the loop.

\subsection{Regulating Search through Monitoring and Control}
\label{sec:metacognitive-regulation}
\label{sec:monitoring-control}

Inspired by the monitoring--control framework of Nelson and Narens~\cite{nelsonnarens1990metamemory} and its extension to reasoning and problem solving~\cite{ackermanthompson2017metareasoning}, \projectname{} uses monitoring to update the Cognitive State from search progress and feedback, and control to decide what to investigate next and how. The resulting decisions guide object-level search and, through the research harness, execution and validation.

\subsubsection{Monitoring}
\label{sec:monitoring}

When targets, ideas, evaluations, or evidence change, monitoring reads the current Cognitive State together with relevant object-level judgments, evaluations, and evidence. It updates the selective representation of scientific understanding and assesses what the changes imply for subsequent search, such as whether current support is sufficient or further work would repeat an unresolved limitation. This provides ongoing verification of evidential support and assumptions, with the assessments recorded in the Cognitive State. Applicable operators from the repertoire support these updates:

\begin{equation}
C_{t+1}=\operatorname{Monitor}(q,X_t,C_t;\mathcal{R}).
\label{eq:monitoring-update}
\end{equation}

Updates retain sources, assumptions, applicability conditions, and unresolved alternatives, keeping working hypotheses distinct from external evidence. When no revision is warranted, $C_{t+1}=C_t$.

\subsubsection{Control}
\label{sec:control}

Inspired by work on strategy selection as metareasoning~\cite{liedergriffiths2017strategy}, control combines the Cognitive State's assessments with the repertoire's applicability conditions to decide which target or idea to address, whether to continue, revise, redirect, or stop a branch, and which methods to use. For scientific ideation, method selection and the subsequent object-level search update are summarized as:

\begin{equation}
\mathcal{R}_t=\operatorname{Select}(q,X_t,C_{t+1};\mathcal{R}),\qquad X_{t+1}=\operatorname{Search}(q,X_t,C_{t+1};\mathcal{R}_t).
\label{eq:control-update}
\end{equation}

Select formalizes method selection within meta-level control; Search denotes the subsequent object-level application of the selected entries to the chosen target or idea, retaining changes and their rationale in the Target Tree, Idea Graph, and associated evaluations. Decisions that call for execution or validation are carried out through the harness (Section~\ref{sec:real-world-feedback-method}); the requested output, stopping rule, and budget govern task termination.

For example, the Cognitive State may indicate that several ideas share an untested premise, so further development along those branches could preserve a common failure condition. Control therefore selects Assumption and boundary audit and Discriminative prediction and calls for a subtarget to examine the premise. Object-level search makes the premise explicit and derives predictions that distinguish whether it holds, providing a basis for validation. Validation results enter the Evidence State and return to monitoring, which revises the state to inform subsequent decisions.

\begin{algorithm}[H]
    \caption{Scientific Discovery through Monitoring and Control in \projectname{}}
    \label{alg:taiyi-discovery}
    \small
    \setlength{\parskip}{0pt}
    \algrenewcommand{\algorithmicrequire}{\textbf{Input:}}
    \algrenewcommand{\algorithmicensure}{\textbf{Output:}}
    \begin{algorithmic}[1]
        \Require Research question $q$, evidence $E_0$, Cognitive Repertoire $\mathcal{R}$, stopping rule, and budget
        \Ensure Research outputs with supporting evidence and unresolved questions
        \State Initialize $T_0$ with root question $q$, $G_0$ with no ideas, and $C_0$ from $q,E_0$
        \State Set $X_0=(T_0,G_0,E_0)$ and $t\gets0$
        \While{the stopping rule is unmet and resources remain}
            \State $C_{t+1}\gets\operatorname{Monitor}(q,X_t,C_t;\mathcal{R})$
            \State Control chooses the next research task and focus using $q$, $X_t$, and $C_{t+1}$
            \If{the task requires scientific ideation}
                \State $\mathcal{R}_t\gets\operatorname{Select}(q,X_t,C_{t+1};\mathcal{R})$
                \State $X_{t+1}\gets\operatorname{Search}(q,X_t,C_{t+1};\mathcal{R}_t)$
                \State Retain target and idea updates, evaluations, and lineage
            \Else
                \State Refine the task into an execution or validation plan; select tools and skills
                \State Execute the planned task and collect results and feedback
                \State Form $X_{t+1}$ by linking results to targets and ideas; preserve evidence in $E_{t+1}$
            \EndIf
            \State Assess progress against the requested output; set $t\gets t+1$
        \EndWhile
        \State Monitor final outcomes and assess outputs against evidence and applicability conditions
        \State \Return outputs, supporting evidence, and unresolved questions
    \end{algorithmic}
\end{algorithm}

\subsection{Grounding Scientific Ideation in Real-World Feedback}
\label{sec:real-world-feedback-method}

\begin{figure}[H]
    \centering
    \includegraphics[width=0.95\linewidth]{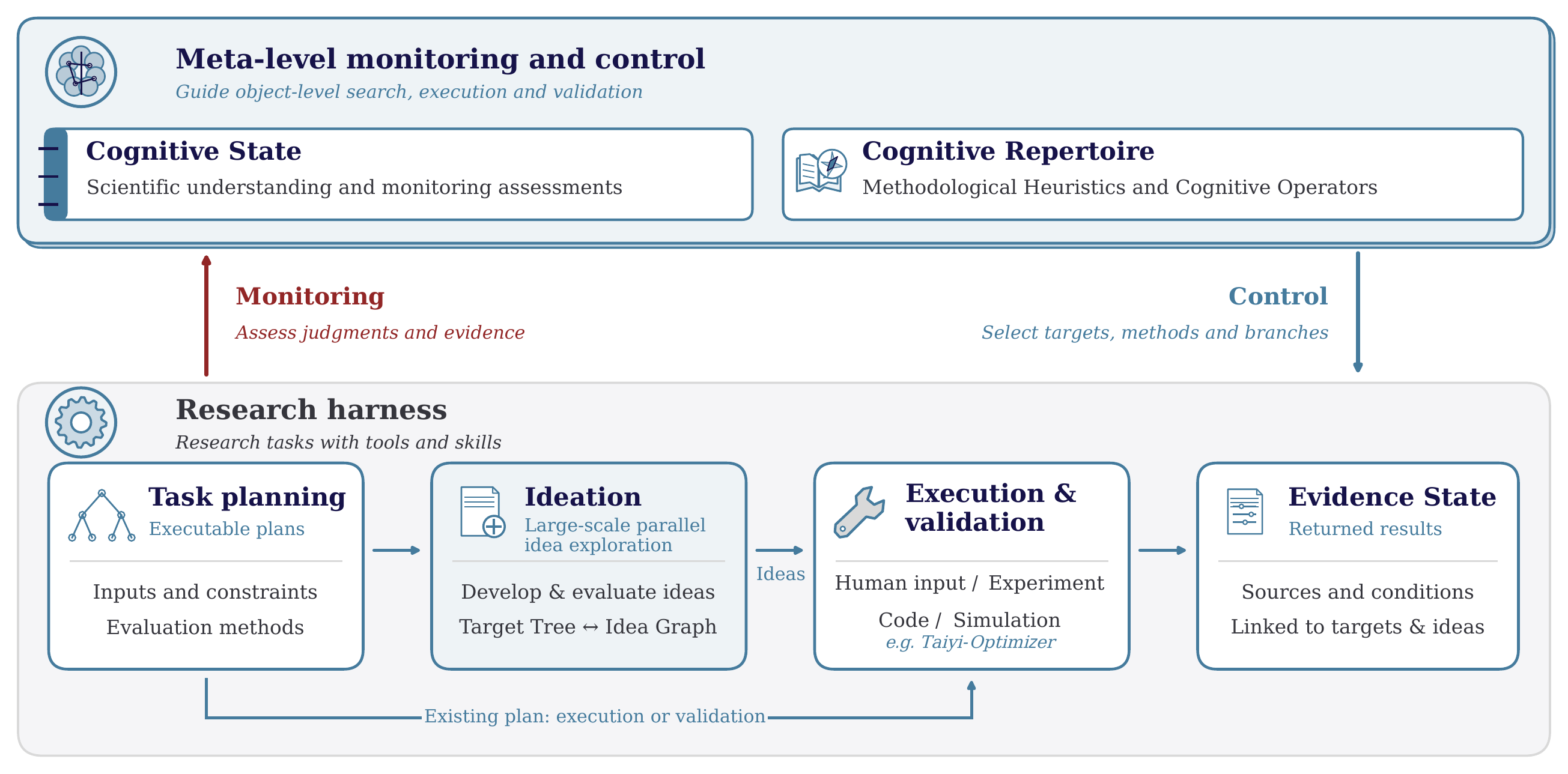}
    \captionsetup{font=small}
    \caption{\textbf{Monitoring and control across research execution.} Control uses the Cognitive State and Cognitive Repertoire to guide research tasks, method selection, and branch decisions. The harness invokes ideation when a new explanation or design is needed; existing plans can proceed directly to execution or validation. Results enter the Evidence State, while monitoring reads relevant object-level judgments, evaluations, and evidence together with the current Cognitive State to update it. Physical experiments and expert participation follow the corresponding research workflow.}
    \label{fig:metacognitive-harness}
\end{figure}

The research harness translates control decisions into research tasks, invoking ideation when a new explanation or design is needed and otherwise proceeding to execution or validation. It refines the selected task into an executable plan with specified inputs, constraints, and evaluation methods, and selects appropriate tools and skills (Figure~\ref{fig:metacognitive-harness}).

During ideation, the Target Tree and Idea Graph guide subagents to explore multiple promising reasoning paths in parallel. Returned results are linked to relevant targets, ideas, and conditions in the Evidence State. Monitoring uses this evidence to verify intermediate conclusions and update the Cognitive State, which then guides control over subsequent search. This combination of parallel exploration and ongoing verification aligns with the Agentic Orchestration and Agentic Verification capabilities of Sage-A1.5~\cite{lou2026sagea15longhorizonexecutionscientific}.

In our research collaborations, this loop connects \projectname{}'s candidate solutions and search formulations to experiments and simulations, with domain experts participating in idea selection and validation. Tools such as \textit{Taiyi-Optimizer} support searches within the specified condition or design spaces, while measurements and simulation results inform subsequent research decisions. Detailed experimental results are presented in Section~\ref{sec:experiments}. Algorithm~\ref{alg:taiyi-discovery} summarizes the logical loop; a task ends with reviewed ideas or executable plans, or continues to execution and validation, according to its requested output and resource budget.

%% file: section/experiments.tex
\newcommand{\systemlogo}[1]{\makebox[1.2em][c]{\raisebox{-0.15em}{\includegraphics[width=0.95em,height=0.95em,keepaspectratio]{figures/logo_#1}}}\hspace{0.25em}}

\section{Experiments}
\label{sec:experiments}

We study \projectname{} through benchmark comparisons and real-world research applications. Taiyi-Bench assesses its scientific ideation capabilities as an AI co-scientist at different knowledge states, through historical rediscovery and open-ended evaluation of idea quality and diversity (Section~\ref{sec:taiyi-bench}). We also evaluate ideation quality and the use of prior research on AgentIdeaBench and IG-Arena (Section~\ref{sec:external-benchmarks}). We then present five applications in \textit{metal catalysis, photochemistry, metamaterials, alloy design, and compressor engineering}, showing how \projectname{} helps address practical scientific and engineering problems (Section~\ref{sec:real-world}).

\subsection{Evaluating Scientific Ideation with Taiyi-Bench}
\label{sec:taiyi-bench}

\subsubsection{Experiment Setup}
\label{sec:experiment-setup}

\paragraph{Taiyi-Bench.}
Taiyi-Bench examines AI co-scientists' ability to propose complete, testable research routes given a scientist's current knowledge. It reconstructs key scientific decisions and their prerequisites from recent published studies\footnote{We select studies first publicly disclosed on or after 1 March 2026, after GPT-5.6-Sol's reported knowledge cutoff (16 February 2026; \href{https://developers.openai.com/api/docs/models/gpt-5.6-sol}{OpenAI documentation}) to reduce training-data contamination risk.} as directed acyclic graphs (DAGs), then captures nontrivial decision points central to each study’s key innovations, and creates tasks that progressively disclose established decisions.Nodes are selected to capture nontrivial decision points central to each study's key innovations. Tasks are constructed by progressively disclosing established decisions. At $L_0$, systems identify the root scientific problem from broad context; subsequent levels specify the problem and disclose preceding decisions. Our evaluation covers 50 papers and 265 case--level tasks across 10 research domains (\textit{i.e.}, earth science, neuroscience, economics, management, physics, biology, chemistry, environmental science, materials science, and communication engineering). Each task requires $K$ self-contained ideas with supporting rationales and discriminating tests. Requesting multiple ideas reflects researchers' need to compare alternative routes before committing experimental resources and tests whether a system can offer substantively different mechanisms rather than variants of one solution.

\paragraph{Metrics.}
Evaluation combines historical rediscovery and open-ended evaluation. For \methodterm{Rediscovery}, \textbf{(1)}~\textbf{Finding} measures identification of the reference root problem at $L_0$. At subsequent levels, \textbf{(2)}~\textbf{Exact} tests whether one proposal completely recovers the reference decisions evaluated at the given information level. \textbf{(3)}~\textbf{Coverage} measures the fraction recovered by the proposal set, while \textbf{(4)}~\textbf{Lookahead} measures the largest fraction of eligible target and downstream decisions recovered by an individual proposal. Recovery requires necessary prerequisites to be disclosed or established within the same proposal. For \methodterm{Open-ended Evaluation}, \textbf{(5)}~\textbf{Novelty} assesses scientific content beyond prior literature. \textbf{(6)}~\textbf{Effectiveness} assesses whether the proposed mechanism and discriminating tests address the stated research gap. \textbf{(7)}~\textbf{Feasibility} assesses necessary capabilities, scientific constraints, and design validity. \textbf{(8)}~\textbf{Diversity} measures distinct relevant mechanism classes across the proposed ideas. A literature-grounded verifier applies decision-level rubrics and mechanism-equivalence criteria to score these dimensions. Its assessments show good agreement with human expert judgments. Scores are reported on a 0--100 scale, with higher values indicating better performance. The Rediscovery and Open-ended scores each average their four constituent metrics. Overall is the mean of these two scores. We also report \textbf{ELO} ratings~\cite{elo1978rating} from a Bradley--Terry model~\cite{bradley1952rank} of task-level pairwise outcomes. 

\paragraph{Baselines.}
We compare \projectname{} with ten baseline configurations in three groups: \textbf{(1) Direct generation}, comprising GPT-5.6-Sol~\cite{openai_models}, Claude-Opus-5~\cite{anthropic_claude}, and Gemini-3.7-Flash~\cite{google_gemini}; \textbf{(2) General-purpose harnesses}, comprising Codex~\cite{openai_codex} with GPT-5.6-Sol at high and ultra reasoning settings, and Claude Code~\cite{anthropic_claudecode} with GPT-5.6-Sol or Claude-Opus-5; and \textbf{(3) Scientific discovery harnesses}, comprising EvoScientist~\cite{lyu2026evoscientist}, AI Scientist-v2~\cite{yamada2025aiscientistv2}, and Robin~\cite{ghareeb2026robin}. Since this comparison focuses on scientific ideation, we retain the ideation stages of the scientific discovery harnesses and omit their subsequent experiment-execution and report-generation stages.

\paragraph{Implementation Details.}
\projectname{} uses GPT-5.6-Sol at high reasoning effort to facilitate comparison with baseline harnesses using the same base model and reasoning setting. Each system is evaluated on $K=10$ ideas per task. Base models and reasoning settings are summarized in Table~\ref{tab:system-config}. 

\begin{table}[H]
    \centering
    \footnotesize
    \setlength{\tabcolsep}{3pt}
    \renewcommand{\arraystretch}{1.25}
    \caption{Configurations used in the Taiyi-Bench evaluation. Parentheses indicate reasoning effort; ``/'' denotes direct generation without a harness.}
    \label{tab:system-config}
    \begin{tikzpicture}
    \node[inner sep=0pt, outer sep=0pt] (systemconfig) {
    \begin{tabularx}{\linewidth}{@{}p{0.09\linewidth}p{0.13\linewidth}Y@{\hspace{1.5em}}p{0.09\linewidth}p{0.13\linewidth}Y@{}}
        \toprule
        Category & Harness & Model  & Category & Harness & Model  \\
        \cmidrule(r){1-3}\cmidrule(l){4-6}
        \multirow{3}{*}{\makecell[l]{Direct\\generation}}
            & / & GPT-5.6-Sol (high)
            & \multirow{4}{*}{\makecell[l]{General-\\purpose\\harnesses}} & Codex & GPT-5.6-Sol (ultra) \\
            & / & Claude-Opus-5 (high)
            & & Codex & GPT-5.6-Sol (high) \\
            & / & Gemini-3.7-Flash (high)
            & & Claude Code & GPT-5.6-Sol (high) \\
        \noalign{\vbox to 0pt{\vss\hbox{\rule{\dimexpr0.5\linewidth-0.75em\relax}{0.3pt}}\vss}}
        \multirow{3}{*}{\makecell[l]{Scientific\\discovery\\harnesses}}
            & EvoScientist & GPT-5.6-Sol (high)
            & & Claude Code & Claude-Opus-5 (high) \\
        \noalign{\vbox to 0pt{\kern-0.5pt\hbox{\hspace{\dimexpr0.5\linewidth+0.75em\relax}\rule{\dimexpr0.5\linewidth-0.75em\relax}{0.3pt}}\vss}}
            & AI Scientist-v2 & GPT-5.6-Sol (high)
            & \cellcolor{methodlabelfill}Ours & \cellcolor{methodlabelfill}\textbf{\projectname{}} & \cellcolor{methodlabelfill}\textbf{GPT-5.6-Sol (high)} \\
            & Robin & GPT-5.6-Sol (high)
            & & & \\
        \bottomrule
    \end{tabularx}};
    \end{tikzpicture}
\end{table}

\subsubsection{Experimental Results}
\label{sec:main-result}

\paragraph{Leading performance across all metrics.}
\projectname{} achieves the highest Overall score of 54.0, exceeding the strongest baseline, AI Scientist-v2, by 6.1 points (Table~\ref{tab:main-results}). It ranks first across all eight metrics, with Rediscovery and Open-ended scores of 44.6 and 63.3, respectively.

\begin{table}[H]
    \centering
    \scriptsize
    \setlength{\tabcolsep}{1.7pt}
    \caption{Taiyi-Bench results across 50 articles and 265 case--level tasks. Finding uses $L_0$; the other seven metrics use subsequent levels. Avg. (average) denotes the mean of the four metrics within each evaluation group; Overall averages the two group scores. Scores are reported on a 0--100 scale with two decimal places. Bold and underlining mark the \textbf{highest} and \underline{second-highest} unrounded values in each column, respectively. Abbreviations: Find., Finding; Cover., Coverage; Look., Lookahead; Nov., Novelty; Effect., Effectiveness; Feas., Feasibility; Div., Diversity.}
    \label{tab:main-results}
    \begin{tabularx}{\linewidth}{@{}Y*{11}{c}@{}}
        \toprule
        \multirow{2}{*}{System} & \multicolumn{5}{c}{Rediscovery} & \multicolumn{5}{c}{Open-ended} & \multirow{2}{*}{Overall} \\
        \cmidrule(lr){2-6}\cmidrule(lr){7-11}
        & Find. & Exact & Cover. & Look. & Avg. & Nov. & Effect. & Feas. & Div. & Avg. & \\
        \midrule
        \rowcolor{black!6}
        \multicolumn{12}{@{}l}{\textit{Direct generation}} \\
        \systemlogo{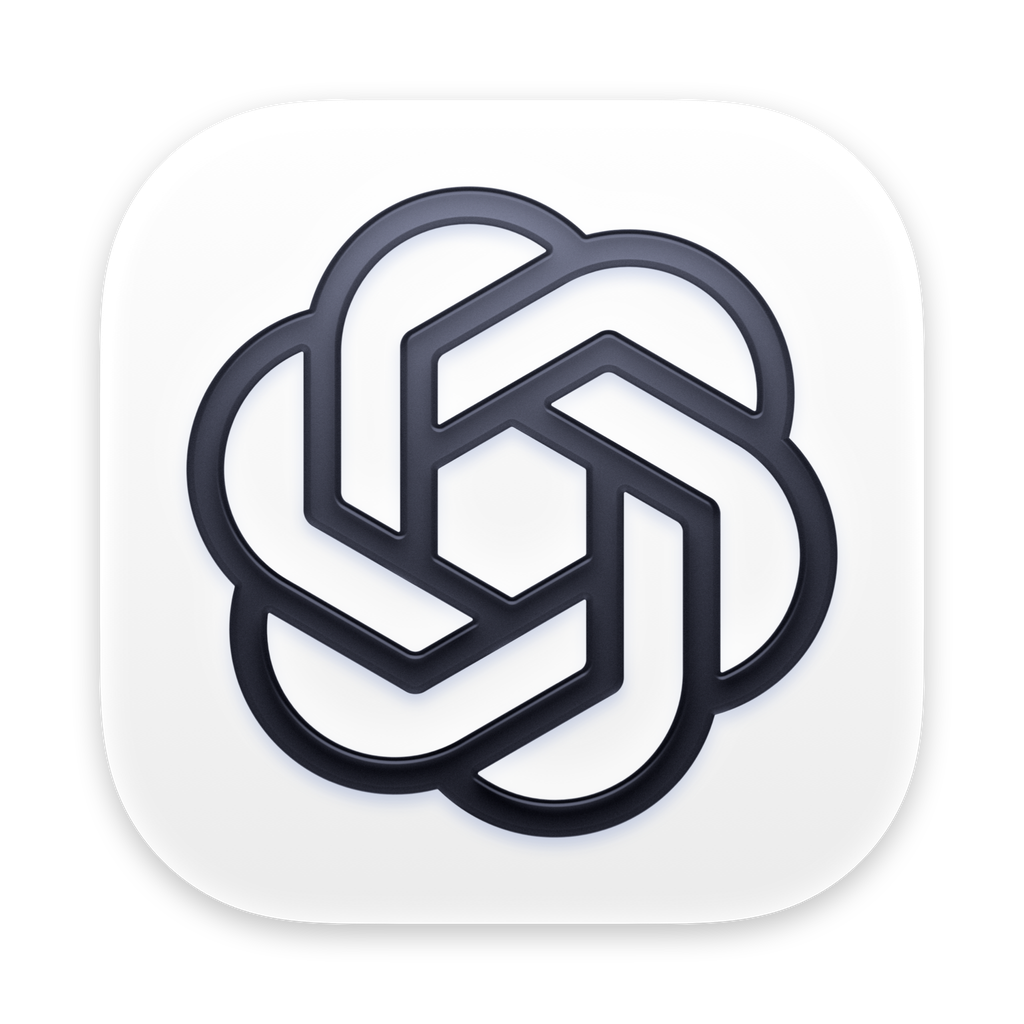}GPT-5.6-Sol & 72.48 & 9.30 & 25.27 & 23.28 & 32.58 & 39.17 & 66.02 & 49.32 & 31.63 & 46.54 & 39.56 \\
        \systemlogo{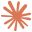}Claude-Opus-5 & 72.32 & 8.02 & 23.35 & 21.34 & 31.26 & 42.66 & 58.18 & 45.78 & 28.70 & 43.83 & 37.54 \\
        \systemlogo{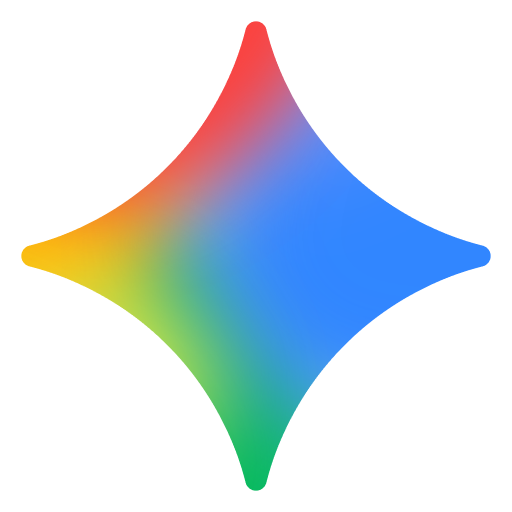}Gemini-3.7-Flash & 66.18 & 7.44 & 17.90 & 17.59 & 27.28 & 41.01 & 56.05 & 41.13 & 28.89 & 41.77 & 34.53 \\
        \midrule
        \rowcolor{black!6}
        \multicolumn{12}{@{}l}{\textit{General-purpose harnesses}} \\
        \systemlogo{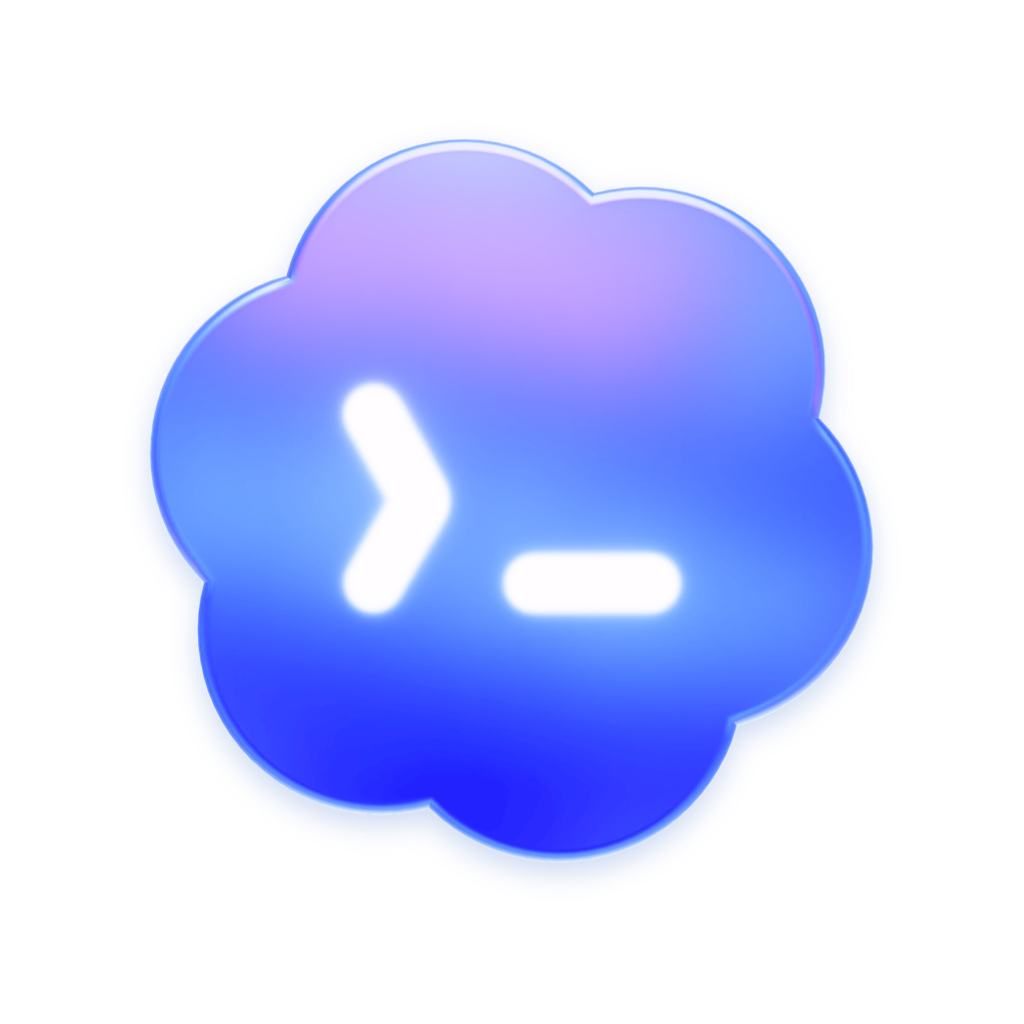}\mbox{Codex} \mbox{(GPT-5.6-Sol, ultra)} & 74.68 & 11.63 & 30.18 & 27.22 & 35.93 & 43.74 & 67.88 & 49.99 & 32.11 & 48.43 & 42.18 \\
        \systemlogo{codex.png}\mbox{Codex} \mbox{(GPT-5.6-Sol)} & 71.86 & 9.30 & 25.16 & 23.30 & 32.41 & 39.82 & 65.81 & 49.55 & 30.51 & 46.42 & 39.41 \\
        \systemlogo{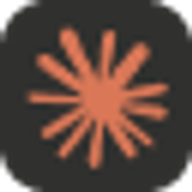}\mbox{Claude Code} \mbox{(GPT-5.6-Sol)} & \underline{78.30} & 13.95 & \underline{35.87} & \underline{33.04} & \underline{40.29} & 43.10 & 67.56 & 50.21 & 31.07 & 47.98 & 44.14 \\
        \systemlogo{claude_code.png}\mbox{Claude Code} \mbox{(Claude-Opus-5)} & 70.90 & 9.30 & 23.21 & 21.88 & 31.32 & 43.54 & 56.95 & 43.91 & 27.67 & 43.02 & 37.17 \\
        \midrule
        \rowcolor{black!6}
        \multicolumn{12}{@{}l}{\textit{Scientific discovery harnesses}} \\
        \systemlogo{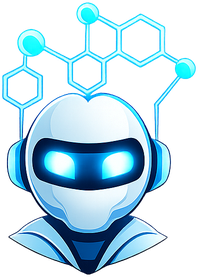}\mbox{EvoScientist} \mbox{(GPT-5.6-Sol)} & 73.62 & 8.88 & 26.70 & 24.23 & 33.36 & 40.92 & 65.33 & \underline{51.05} & 28.06 & 46.34 & 39.85 \\
        \systemlogo{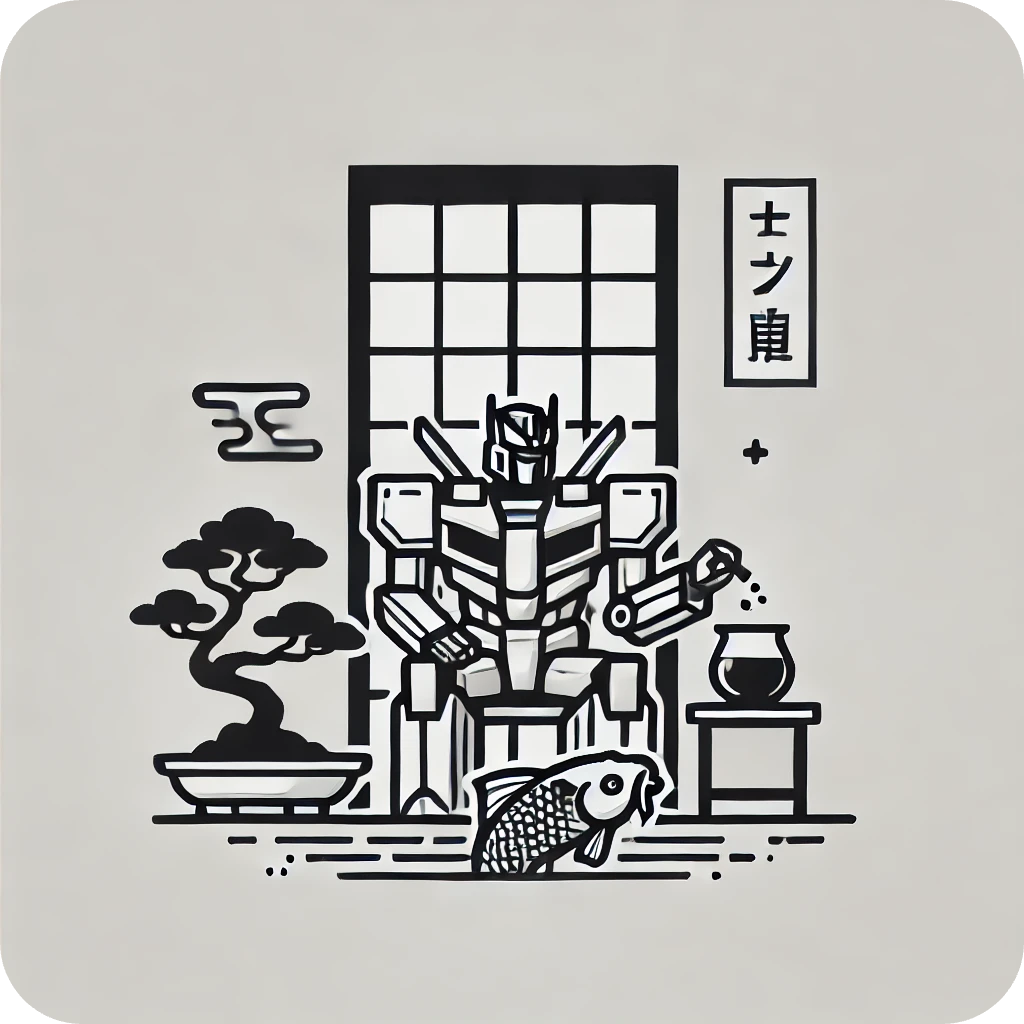}\mbox{AI Scientist-v2} \mbox{(GPT-5.6-Sol)} & 72.54 & \underline{16.74} & 32.37 & 31.62 & 38.32 & \underline{55.20} & \underline{87.23} & 47.73 & \underline{39.40} & \underline{57.39} & \underline{47.85} \\
        \systemlogo{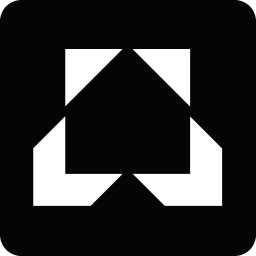}\mbox{Robin} \mbox{(GPT-5.6-Sol)} & 73.31 & 7.91 & 24.88 & 23.07 & 32.29 & 42.17 & 65.26 & 49.90 & 27.47 & 46.20 & 39.24 \\
        \midrule
        \rowcolor{methodlabelfill}
        \systemlogo{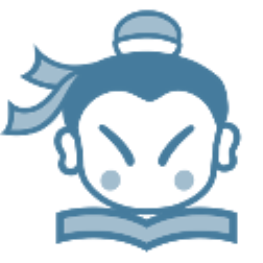}\textbf{\projectname{}} \mbox{(GPT-5.6-Sol)} & \textbf{79.70} & \textbf{19.53} & \textbf{40.69} & \textbf{38.55} & \textbf{44.62} & \textbf{70.54} & \textbf{87.81} & \textbf{52.45} & \textbf{42.33} & \textbf{63.28} & \textbf{53.95} \\
        \bottomrule
    \end{tabularx}
\end{table}

\begin{figure}[htbp]
    \centering
    \includegraphics[width=\linewidth]{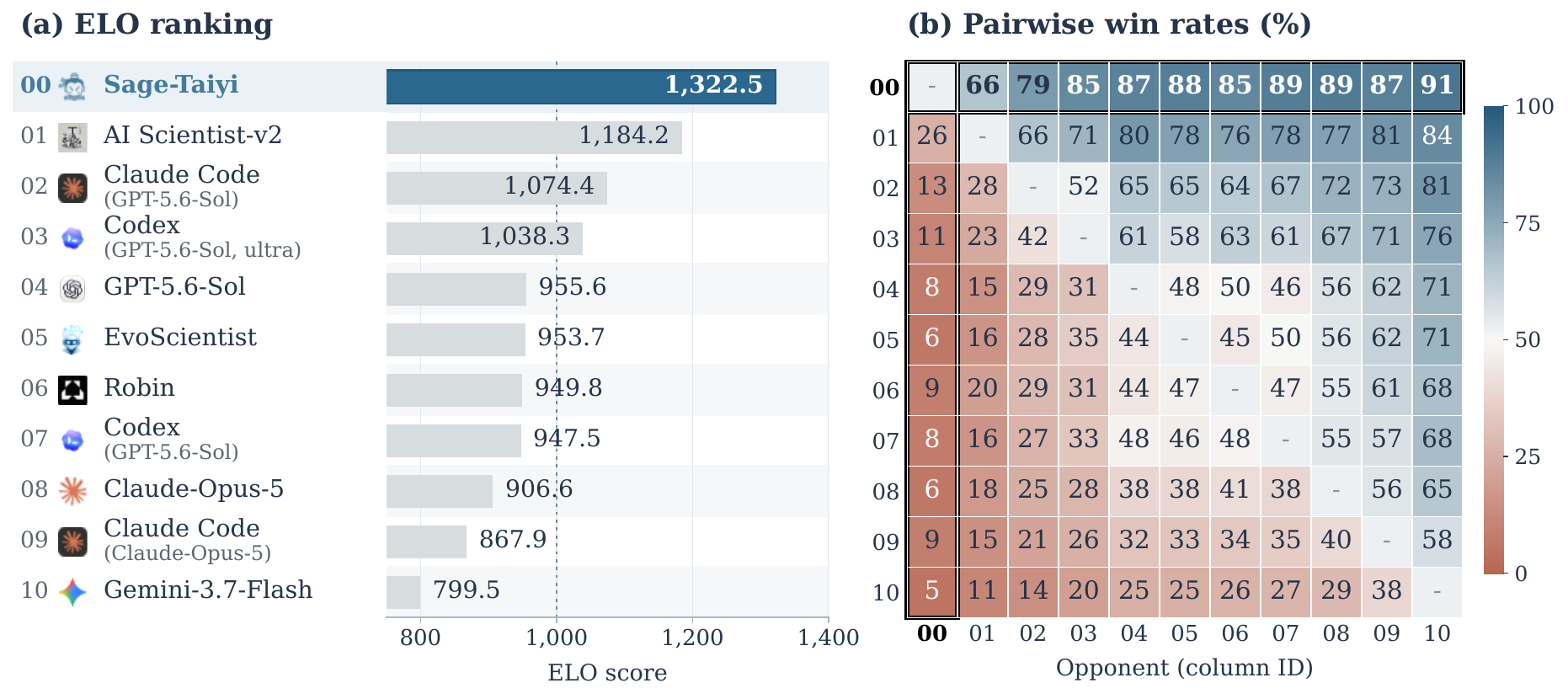}
    \caption{\textbf{ELO ranking and pairwise win rates.}
    \textbf{(a)} ELO scores for all 11 systems. \textbf{(b)} Empirical win percentages of the row system against the column system, with ties retained in the denominator. Both panels use descending ELO order and shared IDs; black outlines mark \projectname{} (00).}
    \label{fig:level-pairwise-winrate}
\end{figure}

Pairwise comparisons support this aggregate ranking: \projectname{} obtains the highest Elo rating of 1,322.5 and wins 66.3\% of comparisons against AI Scientist-v2 (Figure~\ref{fig:level-pairwise-winrate}). These results show advantages in both recovering established scientific decisions and proposing alternative solutions.

\paragraph{Stronger recovery of scientific decisions.}
\projectname{} shows stronger recovery of established scientific decisions from the knowledge disclosed in each task. Its Coverage and Lookahead scores reach 40.7 and 38.6, exceeding the strongest baseline scores of 35.9 and 33.0; its Exact score is also the highest at 19.5 (Table~\ref{tab:main-results}). Because recovered decisions require their prerequisites to be disclosed or established within the same proposal, these results indicate an ability to connect scientific choices with the reasoning needed to support them. Its stronger responses link task constraints, proposed mechanisms, and subsequent design decisions into coherent research routes. Section~\ref{sec:trajectory-analysis} provides a detailed example.

\paragraph{More novel and diverse ideas.}
\projectname{}'s clearest advantage is Novelty: its score of 70.5 exceeds AI Scientist-v2's 55.2 by 27.8\% (Table~\ref{tab:main-results}). Its leading Diversity score of 42.3 further indicates a broader range of mechanism-level alternatives. Qualitatively, its stronger proposals reconsider assumptions or combine complementary mechanisms, explaining how the resulting ideas address the research gap and what predictions could distinguish them. \projectname{} also ranks first in Effectiveness (87.8) and Feasibility (52.4). Together, these results are consistent with \projectname{}'s structured search: the Target Tree tracks scientific requirements, while the Cognitive Repertoire and Idea Graph support the development of diverse alternatives. Monitoring and control use updated assessments to guide subsequent target and method selection.

\begin{figure}[H]
    \centering
    \includegraphics[width=\linewidth]{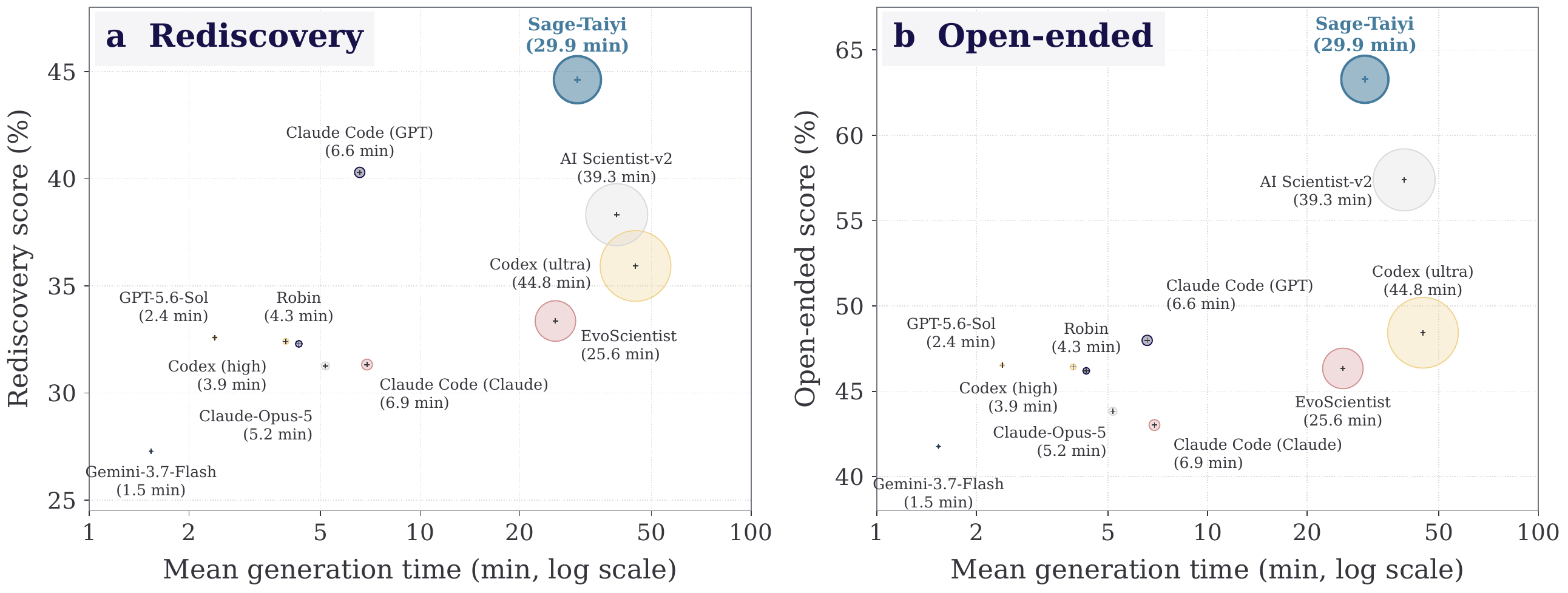}
    \caption{\textbf{Scientific ideation quality and generation time on Taiyi-Bench.} (a) Rediscovery and (b) Open-ended scores, each averaging four metrics. Both panels report mean generation time over all tasks. Crosses mark time–score coordinates; circle diameters scale with time. Generation time is plotted on the horizontal logarithmic axis and score on the vertical linear axis.}
    \label{fig:quality-time}
\end{figure}

\paragraph{Higher quality in less time.}
\projectname{} achieves higher scores in both evaluation groups with 29.9 minutes of recorded generation time, compared with 39.3 minutes for AI Scientist-v2 and 44.8 minutes for Codex (ultra) (Figure~\ref{fig:quality-time}). Its quality advantage therefore accompanies shorter generation times than these two baselines. Parallel exploration of independent research branches allows diverse alternatives to be developed concurrently, helping limit wall-clock generation time. Direct generation and several other harnesses finish faster but achieve lower scores.

\begin{figure}[H]
    \centering
    \includegraphics[width=0.8\linewidth]{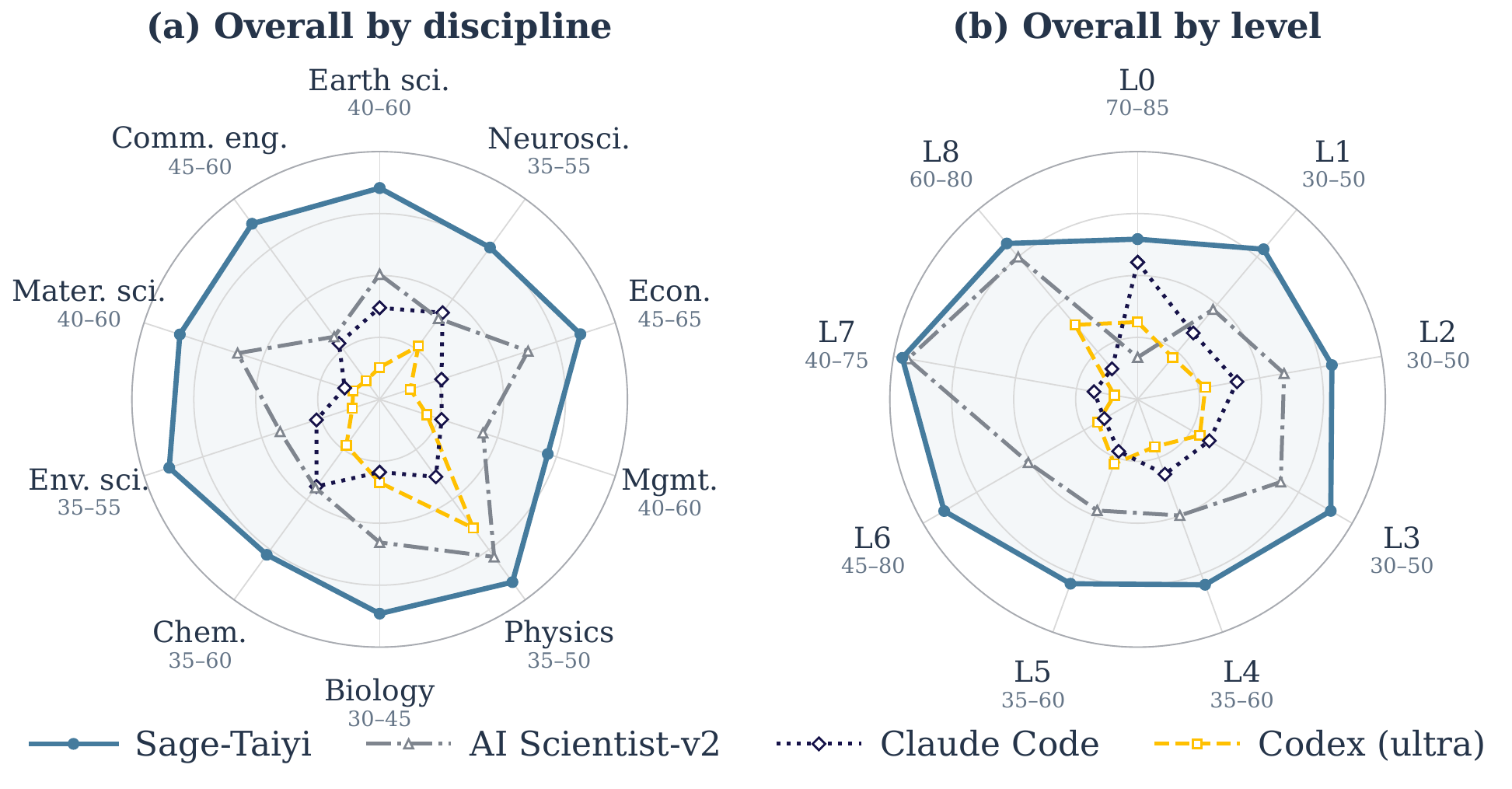}
    \caption{\textbf{Scientific ideation across disciplines and information levels.}
    \textbf{(a)} Overall across ten disciplines, calculated as the mean of Rediscovery and Open-ended scores. \textbf{(b)} Overall across $L_0$--$L_8$, using Finding at $L_0$ and the seven-metric mean thereafter. Later levels differ in task count and targets. In both panels, the gray labels give each spoke's linear score range from center to rim, shared by all four systems. All four systems use GPT-5.6-Sol.}
    \label{fig:levels-budget}
\end{figure}

\paragraph{Performance across fields and knowledge states.}
Among the four systems shown, \projectname{} achieves the highest Overall score in all ten disciplines, with leads of 1.9 to 9.4 points over the strongest baseline in each discipline (Figure~\ref{fig:levels-budget}a). It also achieves the highest score at every level, from $L_0$ to $L_8$ (Figure~\ref{fig:levels-budget}b).

\subsubsection{Case Study}
\label{sec:trajectory-analysis}

\begin{figure}[p]
    \centering
    \includegraphics[width=\linewidth]{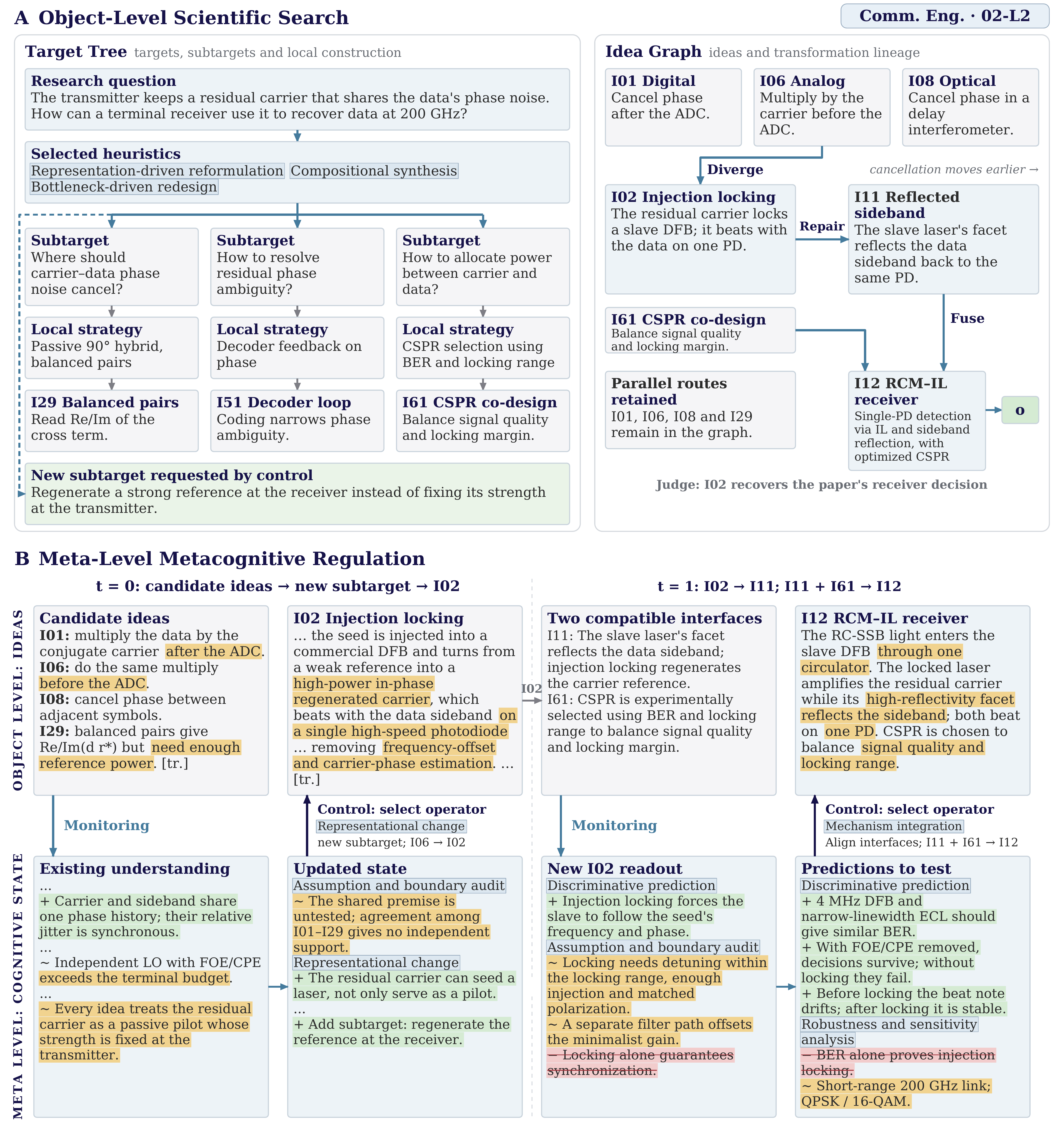}
    \captionsetup{font=normalsize,justification=raggedright,singlelinecheck=false}
    \setlength{\fboxsep}{1pt}
    \caption{\textbf{Scientific search and metacognitive regulation on a communication engineering task ($L_2$).}
    \textbf{(a)} Target Tree and Idea Graph; \textbf{(b)} monitoring and control. I: idea; $o$: output.
    Target/strategy wording, I11/I12, state updates and evolution links are reconstructed; I02 was question-direct and exported as F02.
    \colorbox[HTML]{D5EBD3}{\strut\textcolor{black}{Green ($+$)}}: added understanding;
    \colorbox[HTML]{F1D38F}{\strut\textcolor{black}{Yellow ($\sim$)}}: scope/uncertainty;
    \colorbox[HTML]{F2CCCC}{\strut\textcolor{black}{Red ($-$)}}: excluded inferences, struck through but retained.
    Boxes mark heuristics/operators; blue/navy inter-level arrows mark monitoring/control.
    RCM/RC-SSB: residual-carrier modulation/single-sideband form; IL: injection locking; PD: photodiode; CSPR: carrier-to-signal power ratio; BER: bit-error rate; [tr.]/ellipses: translated/omitted text.}
    \label{fig:cognitive-trajectory}
\end{figure}

We use \projectname{}'s trajectory on a communication engineering task at $L_2$ in Taiyi-Bench to illustrate how its key design components organize scientific exploration (Figure~\ref{fig:cognitive-trajectory}). The question $q$ is \textit{how to recover baseband data in an approximately 200\,GHz link under terminal device, bandwidth, and real-time processing constraints}~\cite{guo2026minimalist}. The decision targeted at $L_1$ is supplied as background at $L_2$: \textit{residual-carrier single-sideband modulation (RC-SSB) retains a same-pump carrier that co-propagates with the data sideband, giving them correlated phase disturbances}. The remaining challenge is \textit{to turn this shared phase history into a stable, economical receiver}. The figure connects object-level search $X_t=(T_t,G_t,E_t)$, the Cognitive State $C_t$, and the Cognitive Repertoire $\mathcal{R}$, highlighting how scientific search and scientific understanding develop together through monitoring and control.

\begingroup
\RaggedRight
\paragraph{Object-Level Scientific Search.}
Three methodological heuristics are selected to guide research paths and target decomposition: \methodterm{\mdseries Representation-driven reformulation}, \methodterm{\mdseries Compositional synthesis}, and \methodterm{\mdseries Bottleneck-driven redesign}. The Target Tree decomposes the receiver problem into \textit{where carrier--data phase noise should cancel, how residual phase ambiguity should be resolved, and how to allocate power between carrier and data} (Figure~\ref{fig:cognitive-trajectory}a). Local strategies lead to \textit{balanced detection of carrier--data cross terms} (I29), \textit{decoder-assisted phase inference} (I51), and \textit{experimental carrier-to-signal power ratio (CSPR) selection using BER and locking range} (I61). Compositional synthesis organizes these complementary branches toward a complete receiver; individual operators perform local transformations along this path. The proposals have retained subidea-composition sources; the displayed subtarget and strategy wording provides a readable reconstruction of those sources.\par
\endgroup

The Idea Graph compares \textit{digital conjugation} (I01), \textit{analog carrier multiplication} (I06), and \textit{differential delay interference} (I08) with \textit{injection-locked carrier regeneration} (I02). I02 \textit{uses the weak residual carrier to seed a slave distributed-feedback laser, then beats the regenerated reference with the data sideband on one photodiode}. The figure uses \methodterm{\mdseries Diverge} to illustrate this change of reference function. Its subsequent \methodterm{\mdseries Repair} adds \textit{the sideband-reflection interface} to I02 (I11), and \methodterm{\mdseries Fuse} combines I11 and I61 into I12: single-PD detection via injection locking and sideband reflection, with CSPR optimized for \textit{signal quality and locking margin}. These links are reconstructed; I02 was generated directly from the question and exported as F02.

\paragraph{Monitoring and Control.}
Monitoring updates the Cognitive State as ideas develop, assessing both the scientific understanding represented by the proposals and its implications for further search. At $t=0$, the comparison concerns \textit{how the initial alternatives use the residual carrier}: digital and analog multiplication use an extracted reference, while balanced detection requires sufficient reference power. \methodterm{\mdseries Assumption and boundary audit} assesses their shared dependence on a passive reference and whether further development would preserve that limitation. At $t=1$, monitoring examines the regenerated-reference route, retaining \textit{sufficient seed power, appropriate detuning, and polarization matching} as applicability conditions and identifying \textit{the cost of a separate filtering path}. These successive updates retain task-specific understanding, unresolved dependencies, and their evidential status in $C_t$.

\enlargethispage{\baselineskip}
Control uses these assessments, together with the applicability conditions of entries in $\mathcal{R}$, to select the next search focus and reasoning methods. At $t=0$, \methodterm{\mdseries Representation-driven reformulation} directs attention to using \textit{the residual carrier as a seed for an actively regenerated reference}, motivating a new subtarget. At $t=1$, control selects \methodterm{\mdseries Mechanism integration} to align the interfaces and operating assumptions of \textit{reference regeneration, sideband reflection, and experimental CSPR selection using BER and locking range} in one receiver candidate. The corresponding target and idea transformations are object-level applications of these decisions. \methodterm{\mdseries Discriminative prediction} and \methodterm{\mdseries Robustness and sensitivity analysis} guide tests involving \textit{broad- versus narrow-linewidth sources, beat-note and relative-phase stability, and removal of frequency-offset and carrier-phase estimation}. This connects assessments of the current search to specific requirements for subsequent idea construction and testing, making the respective roles of monitoring and control concrete.

\FloatBarrier
\subsection{Evaluation on AgentIdeaBench and IG-Arena}
\label{sec:external-benchmarks}

\paragraph{Benchmarks.}
We also evaluate \projectname{}'s scientific ideation capabilities on two recently introduced benchmarks, AgentIdeaBench~\cite{mo2026agentideabench} and IG-Arena~\cite{zhou2026ideagenomes}, and report results using their respective scoring frameworks. AgentIdeaBench combines ratings of originality, impact, feasibility, clarity, and specificity into a weighted score. IG-Arena uses the Population-Evolution Score (PES) to assess how well a proposal builds on prior research and opens a productive direction for follow-up through three dimensions, Heredity, Variation, and Selection.

\paragraph{Evaluation setup.}
We evaluate \projectname{} on 40 AgentIdeaBench subfields with agent-controlled literature search and on all 30 IG-Arena frontier questions across ten scientific domains using Question-setting inputs. In both evaluations, \projectname{} chooses its own search queries and which papers to consult. For IG-Arena, it submits its top-ranked proposal for each question. The benchmark's reference library and lineage graph are withheld during generation.

\begin{table}[H]
    \centering
    \small
    \setlength{\tabcolsep}{4pt}
    \caption{Local AgentIdeaBench results. All systems use GPT-5.6-Sol. The weighted mean uses the dimension weights from Mo et al.~\cite{mo2026agentideabench}. \textbf{Bold} and \underline{underlining} indicate the highest and second-highest scores in each column, respectively.}
    \label{tab:agentideabench}
    \begin{tabularx}{\linewidth}{@{}Y*{6}{c}@{}}
        \toprule
        System & Originality & Impact & Feasibility & Clarity & Specificity & Weighted mean \\
        \midrule
        AI Scientist-v2 & \underline{6.900} & \underline{6.275} & \underline{6.625} & 7.950 & 7.750 & \underline{6.852} \\
        Codex & 6.897 & 6.256 & 6.205 & \underline{8.000} & \underline{7.872} & 6.786 \\
        Claude Code & 6.850 & 6.150 & 5.825 & 7.800 & 7.675 & 6.634 \\
        \rowcolor{methodlabelfill}
        \textbf{\projectname{}} & \textbf{7.300} & \textbf{6.525} & \textbf{7.050} & \textbf{8.100} & \textbf{8.175} & \textbf{7.195} \\
        \bottomrule
    \end{tabularx}
\end{table}

\begin{figure}[H]
    \centering
    \captionsetup{skip=4pt}
    \includegraphics[width=\linewidth]{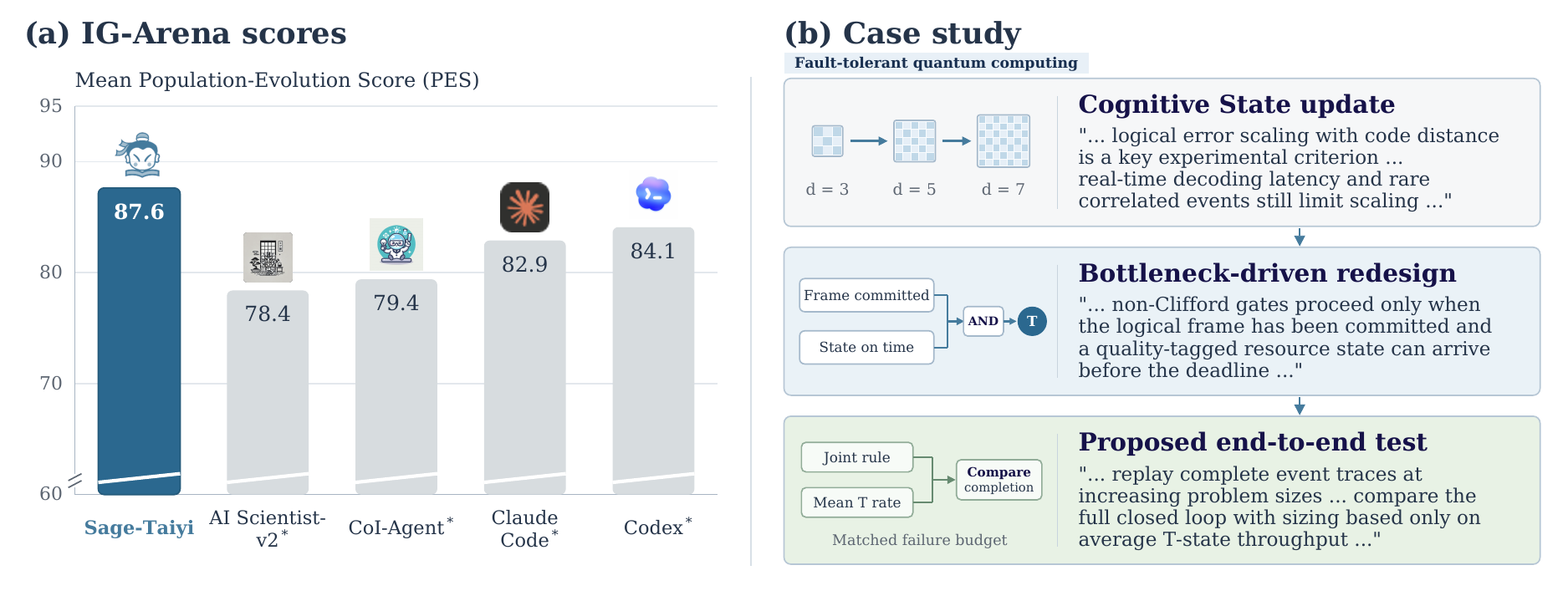}
    \caption{\textbf{IG-Arena results and fault-tolerant quantum computing case.}
    \textbf{(a)} Mean PES on a truncated axis. All systems use GPT-5.5. Asterisks mark published Question-setting scores~\cite[Fig.~6b]{zhou2026ideagenomes}.
    \textbf{(b)} \projectname{} uses surface-code evidence to propose a gate scheduling rule that coordinates decoding and resource-state delivery, with an end-to-end test at a matched failure budget.}
    \label{fig:ig-arena}
\end{figure}

\paragraph{Results.}
On AgentIdeaBench, \projectname{} achieved a weighted mean of 7.195, compared with 6.852 for AI Scientist-v2, the strongest local baseline in the current results (Table~\ref{tab:agentideabench}). \projectname{} had the highest mean on all five dimensions, with originality and feasibility scores 0.400 and 0.425 points higher than AI Scientist-v2, respectively. On IG-Arena, \projectname{} obtained a mean PES of 87.634 in our evaluation. Figure~\ref{fig:ig-arena}a also shows four published reference scores from the Question setting.

\paragraph{Case study on fault-tolerant quantum computing.}
\projectname{} independently retrieved key papers from the IG-Arena reference library (Figure~\ref{fig:ig-arena}b). In updating its Cognitive State, it organized the evidence around a key progression in surface-code research, from observing lower logical error rates at larger code distances to demonstrating sustained error suppression below threshold~\cite{google2023surfacecodescaling,google2025belowthreshold}. It distinguished the evidence for logical memory from the additional requirements of executing a full fault-tolerant algorithm. Bottleneck-driven redesign focused the search on whether real-time decoding and the supply of quantum resource states could both keep pace with gate operations. Compositional synthesis then brought them into a single scheduling rule. A gate that consumes a resource state can proceed only when the relevant decoding result has been committed and the state can arrive by the gate deadline. The proposal specifies an end-to-end comparison with a design sized by average resource-state throughput, testing whether the scheduling rule improves the probability of completing the computation within the same failure budget.

\FloatBarrier
\subsection{Real-World Scientific Discovery}
\label{sec:real-world}

We apply \projectname{} to five real-world research challenges spanning metal catalysis, photochemistry, metamaterials, alloy design, and compressor engineering, translating scientific ideas into new reaction routes, experimentally realized materials, and improved engineering designs. Across these collaborations, \projectname{} helps develop catalytic transformations, overcome restrictive design constraints, and identify effective combinations of composition, processing, and geometry. These advances connect scientific exploration to measurable outcomes, from higher reaction yields and material strength to improved electromagnetic and aerodynamic performance (Table~\ref{tab:real-world}).

Following the research harness described in Section~\ref{sec:real-world-feedback-method}, \projectname{} develops candidate solutions and specifies the variables, objectives, and constraints needed for execution. Domain experts participate in selecting promising ideas and validating their implementation, while Taiyi-Optimizer searches the resulting condition or design spaces. Experimental measurements and simulation results return as evidence, informing subsequent revisions to ideas, conditions, and constraints. This workflow extends \projectname{}'s monitoring--control loop to practical research: experimental and simulation results inform assessments of research progress and guide subsequent decisions about what to investigate and how to proceed.

\begin{table}[H]
    \centering
    \fontsize{10}{13}\selectfont
    \setlength{\parskip}{0pt}
    \setlength{\tabcolsep}{8pt}
    \setlength{\aboverulesep}{5pt}
    \setlength{\belowrulesep}{5pt}
    \captionsetup{skip=7pt}
    \caption{Scientific contributions and outcomes across five real-world research collaborations.}
    \label{tab:real-world}
    \renewcommand{\arraystretch}{1.0}
    \renewcommand{\tabularxcolumn}[1]{m{#1}}
    \begin{tabularx}{\linewidth}{@{}>{\RaggedRight\arraybackslash}m{0.20\linewidth}Y@{}}
        \toprule
        \textbf{Research area} & \textbf{Scientific contribution and outcome} \\
        \midrule
        Metal catalysis & Developed a cobalt-catalysed route to tetrasubstituted allenyl ketones, raising the model-reaction yield from trace amounts to \textbf{84\%} through reaction-condition optimization. \\
        \midrule
        Photochemistry & Developed a photocatalytic route to site-selective allylic amination with heteroaryl amines, increasing the model-reaction yield from \textbf{13\% to 69\%}. \\
        \midrule
        Metamaterials & Designed a multilayer electromagnetic filter and relaxed a restrictive thickness constraint, achieving simulated average transmission of \textbf{6.2\% in the stopband} and \textbf{94.5\% in the passband}. \\
        \midrule
        Alloy design & Jointly designed and optimized magnesium-alloy composition and heat treatment, producing an alloy with a mean measured tensile strength of \textbf{342 MPa}. \\
        \midrule
        Compressor\newline engineering & Developed and refined a centrifugal-compressor design, achieving a CFD-evaluated impeller efficiency of \textbf{91.5\%} at a pressure ratio of \textbf{4.0}. \\
        \bottomrule
    \end{tabularx}
\end{table}

\FloatBarrier
\addtocontents{toc}{\protect\setcounter{tocdepth}{3}}
\subsubsection{Metal Catalysis: Earth-Abundant Metal-Catalyzed Carbonylation}
\label{sec:metal-catalysis}

\textcolor{taiyiblue}{\textbf{Problem statement.}} Unsymmetrical ketones are widespread in pharmaceuticals, agrochemicals and functional materials, and carbonylation with carbon monoxide offers an atom-economical route to assemble them from two different carbon fragments and a C1 unit. The central challenge is chemoselectivity, because the two fragments must be joined at the same carbonyl carbon in a cross-selective manner while the formation of symmetrical ketones is suppressed. Earth-abundant metal catalysts would improve the practicality of this approach, but achieving high cross-selectivity with such catalysts remains difficult when both carbon partners are activated through similar pathways at the same metal centre.

\begin{figure}[!htbp]
    \centering
    \includegraphics[width=\linewidth]{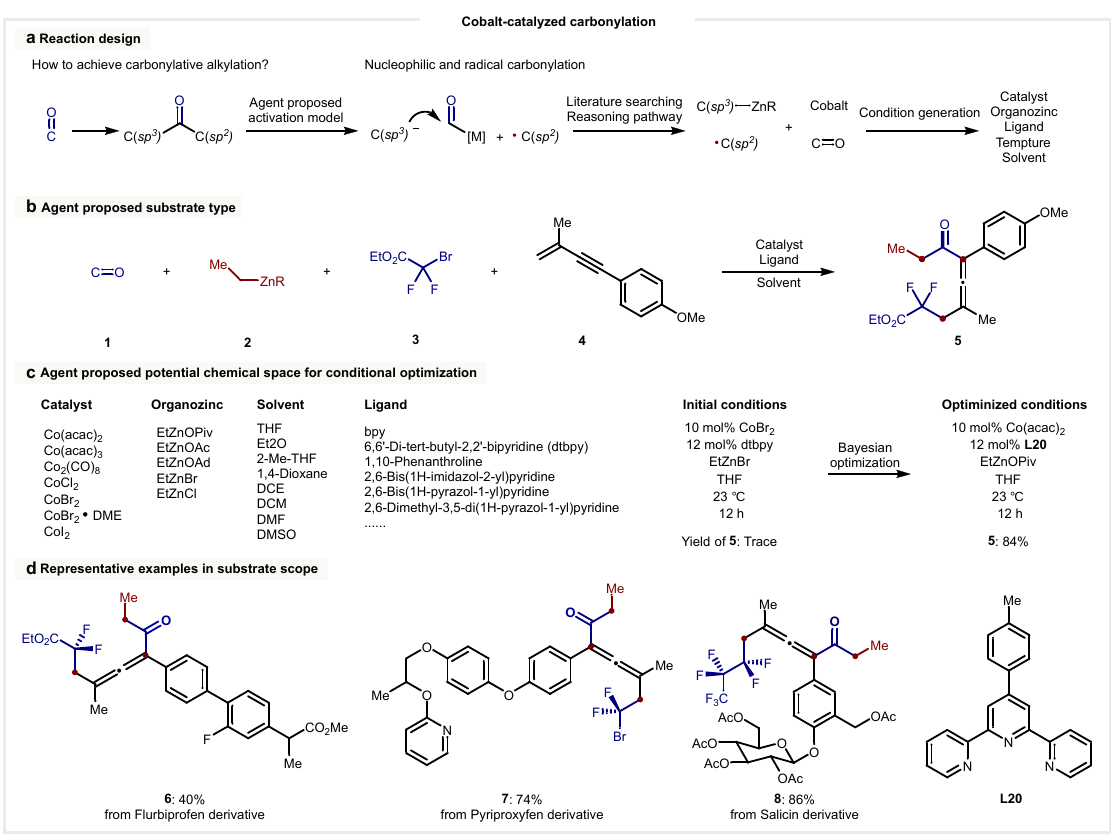}
    \captionsetup{font=small}
    \caption{\textbf{Cobalt-catalysed carbonylation: reaction design and experimental outcomes.} Panels a--d show the activation model, model reaction, condition space and initial/optimized conditions, and bioactive-molecule derivatives with ligand L20. Yields are isolated.}
    \label{fig:metal-catalysis-reaction}
\end{figure}

\begin{figure}[p]
    \centering
    \captionsetup{font=small}
    \includegraphics[width=\linewidth]{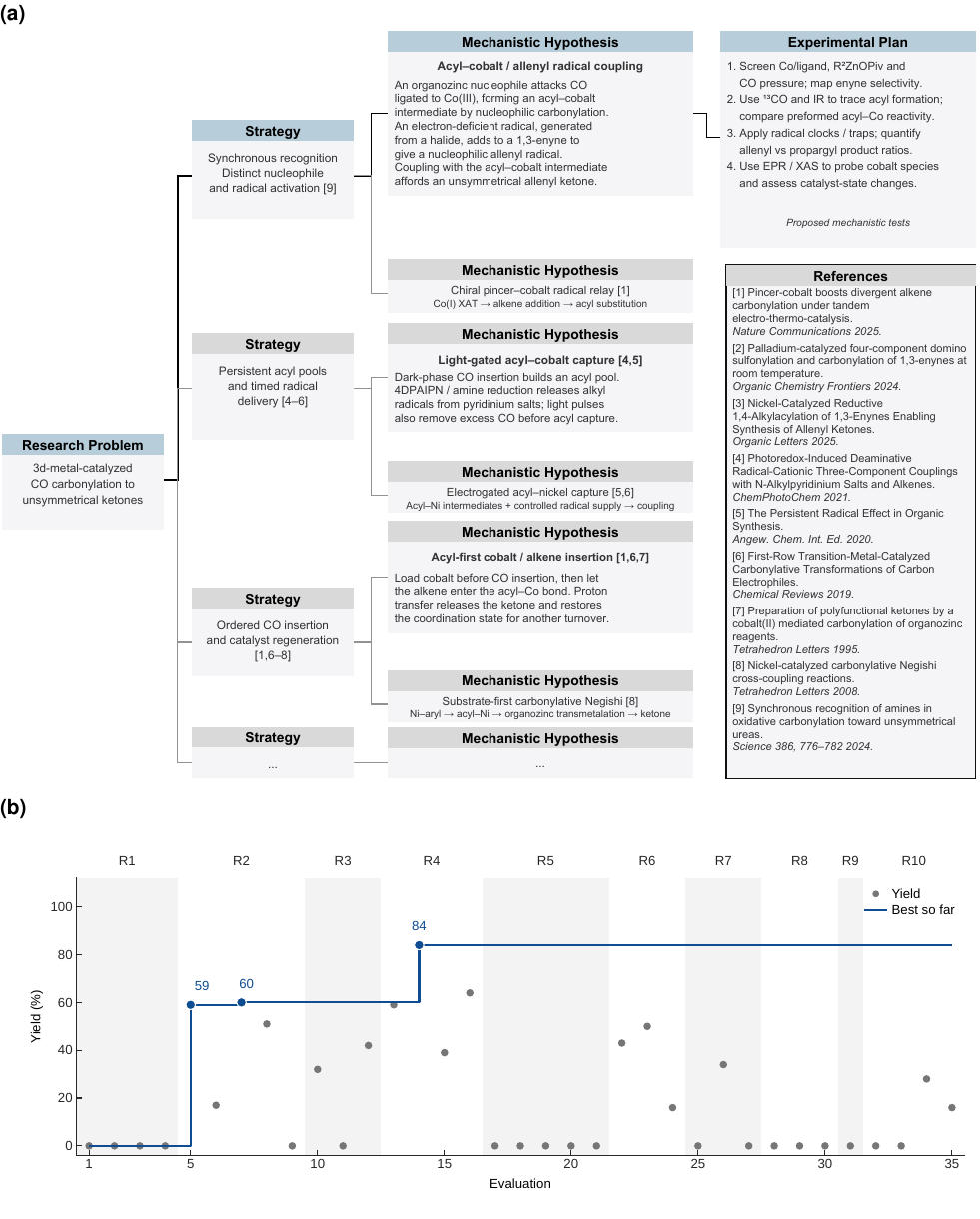}
    \caption{\textbf{Cobalt-catalysed carbonylation: workflow and optimization.} \textbf{(a)} Strategies, mechanistic hypotheses and proposed tests. The highlighted synchronous-recognition route combines nucleophilic carbonylation with polarity reversal and acyl--cobalt/allenyl-radical coupling. Branches group candidate mechanisms; tests remain prospective. Workflow references are numbered locally. \textbf{(b)} Thirty-five evaluations across ten rounds; points show yields and the line tracks the cumulative best, reaching 84\%. The reaction and conditions appear in Figure~\ref{fig:metal-catalysis-reaction}.}
    \label{fig:metal-catalysis}
\end{figure}

\textcolor{taiyiblue}{\textbf{Implementation.}} For this scientific question, \projectname{} proposed several activation strategies (Figure~\ref{fig:metal-catalysis}a). One of them adapts the \textcolor{taiyiblue}{\textbf{synchronous recognition strategy}} previously reported for the synthesis of unsymmetrical ureas~\cite{wang2024synchronous}, in which the two coupling partners are differentiated by distinct activation modes. \projectname{} proposed that metal-mediated carbonylation and radical carbonylation could operate cooperatively at a cobalt centre to achieve this differentiation in ketone synthesis. In the metal-mediated branch, a carbon nucleophile such as an organometallic reagent attacks a CO-ligated Co(III) centre, forming an acyl--cobalt intermediate through nucleophilic carbonylation. In the radical branch, a carbon radical generated from the second coupling partner couples with this intermediate to deliver the unsymmetrical ketone.

To access the carbon radical required in the radical branch, \projectname{} further proposed a \textcolor{taiyiblue}{\textbf{polarity-reversal strategy}}, in which an electron-deficient radical adds to an alkene to generate a nucleophilic alkyl radical. A subsequent literature analysis by \projectname{} showed that 1,3-enynes, which form C(sp$^2$)-centred allenyl radicals upon radical addition, have rarely been employed as radical acceptors in oxidative carbonylation. On this basis, \projectname{} recommended a cobalt-catalysed redox-neutral carbonylation in which organozinc reagents serve as nucleophiles, 1,3-enynes as radical acceptors and electron-deficient halides as radical precursors, affording allenyl ketones (Figure~\ref{fig:metal-catalysis-reaction}). This case illustrates how \projectname{} narrows a general activation strategy to a specific, underexplored reaction design through successive rounds of literature comparison.

\begin{samepage}
Literature searching and mechanistic reasoning mapped the proposed activation model onto an alkylzinc reagent, an allenyl-radical pathway, a cobalt catalyst and CO, and identified the catalyst, organozinc reagent, ligand, solvent and temperature as the variables for condition generation. The resulting model reaction couples CO (1), an ethylzinc reagent (2), ethyl bromodifluoroacetate (3) and the 1,3-enyne 4 to form the tetrasubstituted allenyl ketone 5 (Figure~\ref{fig:metal-catalysis-reaction}, subpanel b). \projectname{} then defined a chemical space for condition optimization comprising seven cobalt sources, five ethylzinc reagents with different counteranions, eight solvents and a series of nitrogen-donor ligands (Figure~\ref{fig:metal-catalysis-reaction}, subpanel c).

\end{samepage}

The resulting condition space was explored using Taiyi-Optimizer through the workflow described in Section~\ref{sec:real-world-feedback-method}, with measured yields from each batch of reactions guiding subsequent condition selection (Figure~\ref{fig:metal-catalysis}b).

\textcolor{taiyiblue}{\textbf{Results.}} Under the initial conditions (10 mol\% CoBr$_2$, 12 mol\% dtbpy, EtZnBr, THF, 23 $^\circ$C, 12 h), only trace amounts of 5 were formed. Bayesian optimization within this space identified Co(acac)$_2$, the terpyridine ligand L20 and EtZnOPiv as the optimal combination, \textcolor{taiyiblue}{\textbf{increasing the yield of 5 to 84\%}} while the solvent, temperature and reaction time remained unchanged. The key changes were the replacement of the bidentate dtbpy with the tridentate L20 and of EtZnBr with EtZnOPiv (Figure~\ref{fig:metal-catalysis-reaction}, subpanel c). The optimization curve reaches the same 84\% yield (Figure~\ref{fig:metal-catalysis}b). The optimized conditions were applicable to enynes derived from bioactive molecules and to different electron-deficient halides, delivering the allenyl ketones 6, 7 and 8 from flurbiprofen, pyriproxyfen and salicin derivatives in 40\%, 74\% and 86\% yield, respectively (Figure~\ref{fig:metal-catalysis-reaction}, subpanel d).

\FloatBarrier
\subsubsection{Photoredox Catalysis: Site-Selective Allylic Amination with Heteroaryl Amines}
\label{sec:photochemistry}

\textcolor{taiyiblue}{\textbf{Problem statement.}} Allylic amines are prevalent in bioactive molecules and serve as versatile intermediates, as the alkene handle allows further elaboration after C--N bond formation. Dehydrogenative C--H/N--H coupling between an alkene and an amine is the most direct route to these motifs, since it avoids prefunctionalization of either partner and releases only hydrogen. The difficulty lies in site selectivity: the alkene presents several allylic C--H bonds of comparable strength, and the intermediate allylic radical can be functionalized at either terminus, so both regio- and positional selectivity must be controlled.

\begin{figure}[H]
    \centering
    \includegraphics[width=0.95\linewidth]{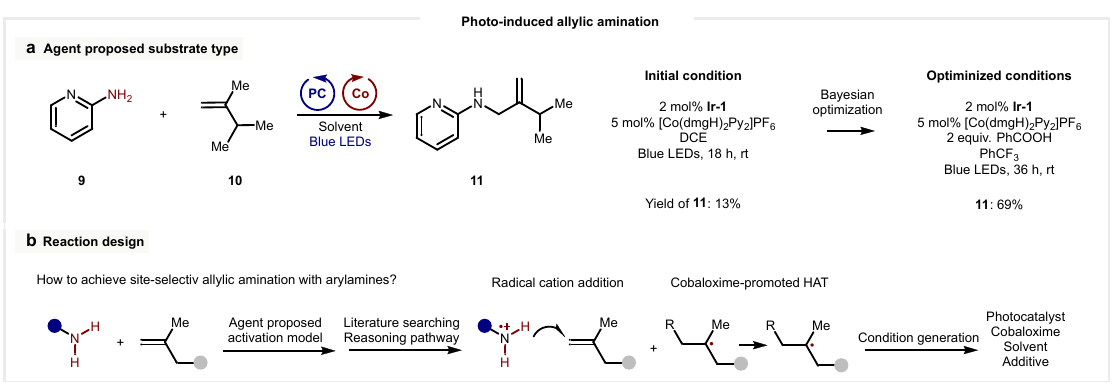}
    \captionsetup{font=small,skip=5pt}
    \caption{\textbf{Photocatalytic allylic amination: reaction design and experimental outcomes.} \textbf{(a)} The 2-aminopyridine model reaction, including initial and optimized conditions. \textbf{(b)} The activation model. PC, photocatalyst; Co, cobaloxime; HAT, hydrogen atom transfer.}
    \label{fig:photochemistry-reaction}
\end{figure}

\begin{figure}[p]
    \centering
    \captionsetup{font=small}
    \includegraphics[width=\linewidth]{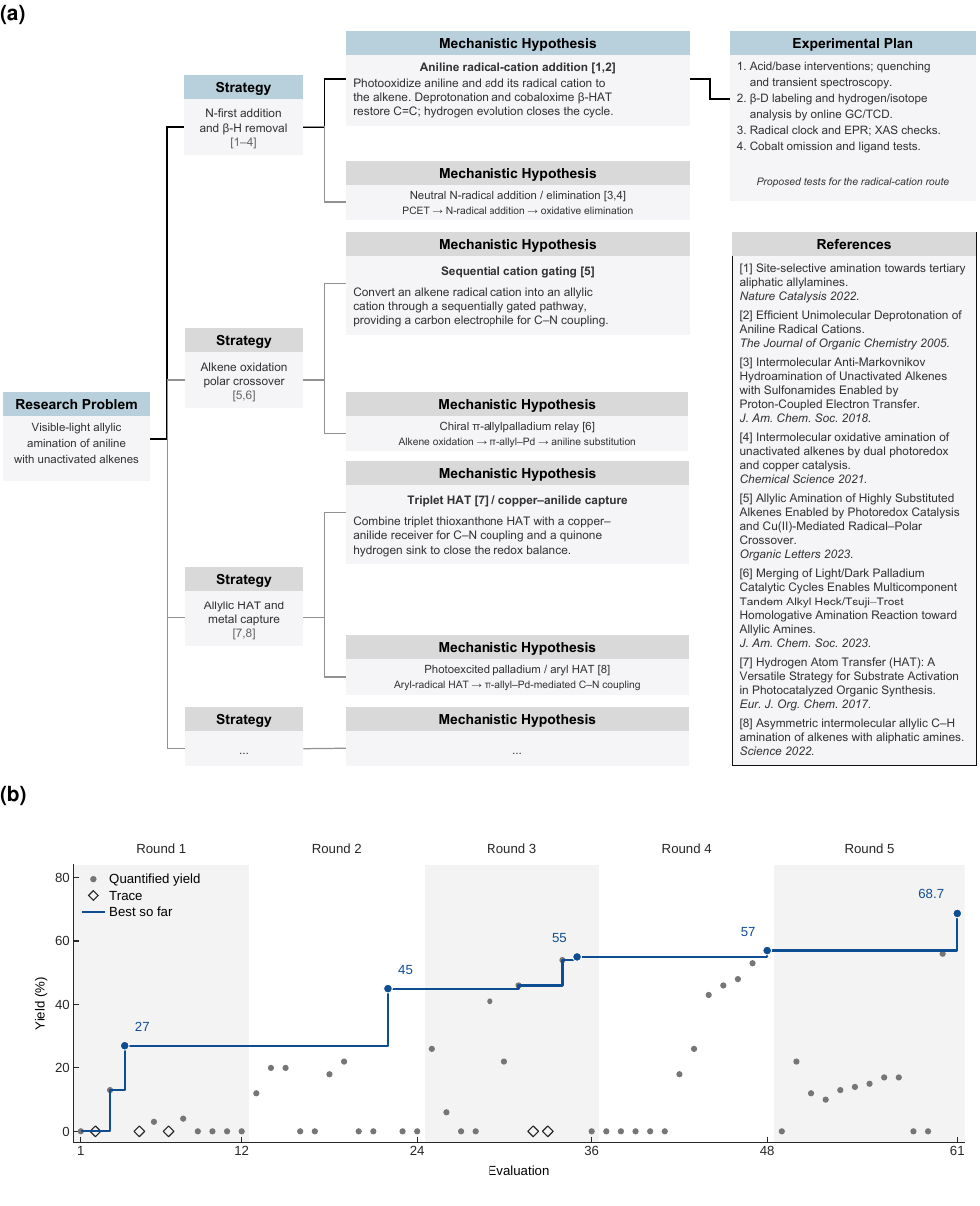}
    \caption{\textbf{Photocatalytic allylic amination: workflow and optimization.} \textbf{(a)} Candidate mechanisms and proposed tests; the highlighted aniline-based route combines radical-cation addition with cobaloxime-mediated HAT. Tests remain prospective. Workflow references are numbered locally; the model reaction uses 2-aminopyridine. \textbf{(b)} Sixty-one evaluations across five rounds; points show quantified yields, open diamonds denote trace amounts, and the line tracks the cumulative best. The maximum, 68.7\%, rounds to the 69\% reported in Figure~\ref{fig:photochemistry-reaction} and the text.}
    \label{fig:photochemistry}
\end{figure}

\begin{samepage}
\textcolor{taiyiblue}{\textbf{Implementation.}} \projectname{} addressed site-selective allylic amination as the scientific question. Its literature analysis showed that photocatalytic dehydrogenative allylic amination has been established for aliphatic amines, whereas the corresponding coupling of alkyl alkenes with anilines and heteroaryl amines has not been reported. \projectname{} attributed this gap to the lower nucleophilicity and higher oxidation potential of aromatic amines, and designed an activation mode in which \textcolor{taiyiblue}{\textbf{single-electron oxidation of the amine generates an aminium radical cation}} that adds to the alkene, followed by cobaloxime-mediated hydrogen atom transfer from the resulting carbon radical to install the alkene and release hydrogen (Figure~\ref{fig:photochemistry}a).

\end{samepage}

Literature searching and mechanistic reasoning identified the photocatalyst, cobaloxime, solvent and additive as the variables for condition generation. The proposed model reaction couples 2-aminopyridine (9) with the 1,1-disubstituted alkene 10 under photoredox and cobalt dual catalysis to give the allylic amine 11 (Figure~\ref{fig:photochemistry-reaction}, subpanel a).

Taiyi-Optimizer explored this condition space using the workflow described in Section~\ref{sec:real-world-feedback-method}. Measured yields from each batch of experiments guided the selection of subsequent photocatalyst, cobaloxime, solvent and additive combinations (Figure~\ref{fig:photochemistry}b).

\textcolor{taiyiblue}{\textbf{Results.}} Under the initial conditions (2 mol\% Ir-1, 5 mol\% [Co(dmgH)$_2$Py$_2$]PF$_6$, DCE, blue LEDs, 18 h, room temperature), 11 was formed in 13\% yield. \textcolor{taiyiblue}{\textbf{Bayesian optimization raised the yield to 69\%}}, with the key changes being a switch of solvent from DCE to PhCF$_3$, the addition of 2 equiv. PhCOOH and an extension of the reaction time to 36 h (Figure~\ref{fig:photochemistry-reaction}, subpanel a). The optimization curve records a best yield of 68.7\%, rounded to 69\% in the reaction scheme and text (Figure~\ref{fig:photochemistry}b).

\FloatBarrier
\subsubsection{Metamaterials: Revising Constraints in Electromagnetic Filter Design}
\label{sec:metamaterials}

\textcolor{taiyiblue}{\textbf{Problem statement.}} The metamaterial collaboration targets a multilayer electromagnetic band-stop filter made from patterned copper on ceramic substrates. The design must suppress transmission over 8--12 GHz while preserving high transmission over 14.5--15.5 GHz. Achieving both objectives requires coordinated choices of structural topology and geometric parameters within fabrication constraints, making the definition of the design space central to the search.

\begin{figure}[p]
    \centering
    \includegraphics[width=\linewidth]{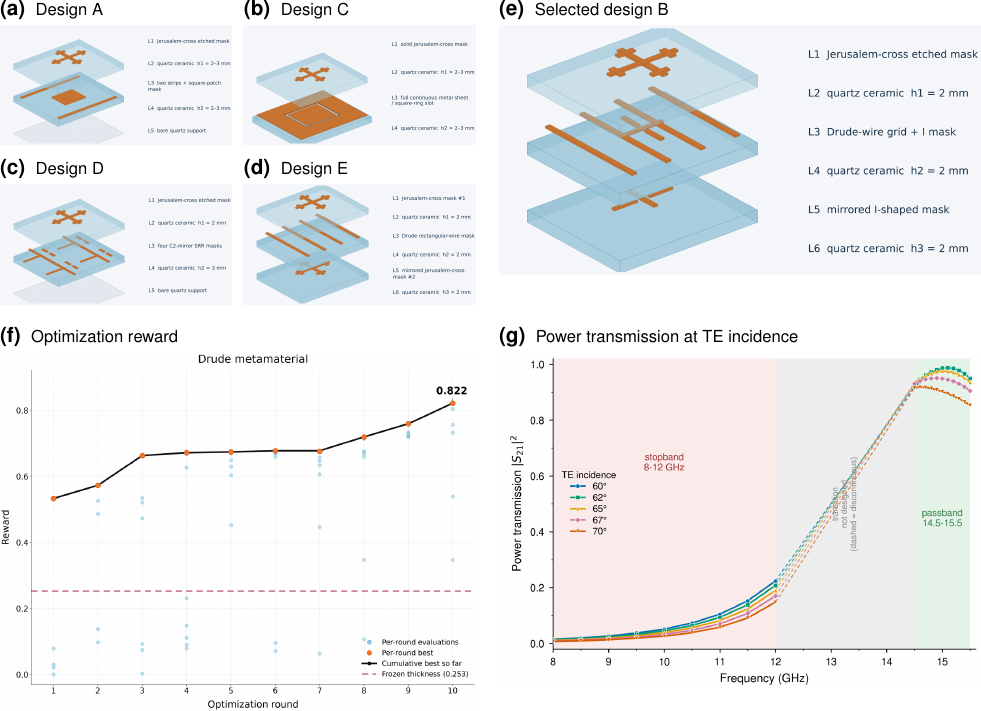}
    \caption{\textbf{Metamaterial topology selection and optimization.}
    \textbf{(a--d)} Alternative designs A, C, D, and E, with source mask labels and nominal ceramic thicknesses: 2--3 mm for A and C, 2 and 3 mm for D, and three 2 mm layers for E.
    \textbf{(e)} Selected design B: a Jerusalem-cross mask, a Drude-wire grid with an I-shaped mask, and a mirrored I-shaped mask interleaved with three quartz-ceramic substrates, each nominally labelled 2 mm; optimized dimensions are given in Table~\ref{tab:metamaterial-results}.
    \textbf{(f)} Ten-round reward history. Light-blue points show per-round evaluations, orange points show per-round best values, and the black line shows the cumulative best. The dashed line marks the frozen-thickness reference of 0.253; the final best reward is 0.822.
    \textbf{(g)} Power transmission $|S_{21}|^2$ for transverse-electric (TE) incidence at $60^\circ$, $62^\circ$, $65^\circ$, $67^\circ$, and $70^\circ$. Shaded target bands are 8--12 and 14.5--15.5 GHz. The intervening region is labelled as not designed, with dashed connections denoting discontinuous spectra.}
    \label{fig:metamaterials}
\end{figure}

\textcolor{taiyiblue}{\textbf{Implementation.}} \projectname{} proposed five configurations: (A) anisotropic strips and patches with an outer Jerusalem cross; (B) Drude wires with an I-shaped Lorentz ceramic-matrix-composite (CMC) half-wave wall and an outer Jerusalem cross; (C) a Babinet-complementary loop-and-slot structure; (D) an in-plane paired four-split-ring-resonator assembly with an outer Jerusalem cross; and (E) mirrored Jerusalem crosses on opposite sides, an embedded Drude wire grid, and a 6 mm half-wave feature. Experts selected design B for electromagnetic simulation and optimization. Figure~\ref{fig:metamaterials}a--d shows the alternatives A, C, D, and E, while panel e shows design B, whose patterned metal masks are interleaved with three quartz-ceramic substrates.

Fabrication considerations initially fixed the substrate thicknesses while the remaining parameters were optimized. The first optimization did not produce satisfactory passband performance. Optimization feedback led \projectname{} to identify fixed substrate thickness as a bottleneck and recommend relaxing this constraint. The team then allowed substrate thickness to vary and repeated the search with Taiyi-Optimizer in the expanded design space. This sequence illustrates the monitoring--control loop described in Section~\ref{sec:monitoring-control}: optimization feedback informs an assessment of the current search constraints, which guides their revision and subsequent exploration.

Following the research harness in Section~\ref{sec:real-world-feedback-method}, Taiyi-Optimizer searched ten continuous geometric variables for the symmetric design. Table~\ref{tab:metamaterial-results} defines the variables and their search ranges. The dielectric-layer thicknesses satisfy $h_1=h_3$, with total thickness $h_{\mathrm{total}}=h_1+h_2+h_3$ and middle-layer fraction $f_{\mathrm{mid}}=h_2/h_{\mathrm{total}}$; $x$ and $y$ are orthogonal in-plane directions. The objective maximizes high passband transmission while penalizing stopband transmission:
\begin{equation}
    R_{\mathrm{meta}}=\overline{T}_{\mathrm{pass}}-1.5\,\overline{T}_{\mathrm{stop}},
\label{eq:metamaterial-reward}
\end{equation}
where $\overline{T}_{\mathrm{pass}}$ and $\overline{T}_{\mathrm{stop}}$ are the dimensionless band averages of power transmission $|S_{21}|^2$, and $S_{21}$ is the forward transmission scattering parameter.

\textcolor{taiyiblue}{\textbf{Results.}} The optimized filter achieved reported average stopband and passband transmissions of 6.2\% and 94.5\%, respectively. Figure~\ref{fig:metamaterials}f shows a frozen-thickness reference reward of 0.253 and a final best reward of 0.822 after ten optimization rounds. The optimized configuration and simulated performance are summarized in Table~\ref{tab:metamaterial-results}.
\begin{table}[H]
    \centering
    \fontsize{10}{12}\selectfont
    \renewcommand{\arraystretch}{1.02}
    \setlength{\tabcolsep}{4pt}
    \setlength{\parskip}{0pt}
    \renewcommand{\tabularxcolumn}[1]{m{#1}}
    \caption{Metamaterial search ranges, optimized configuration, and simulated performance. Ranges specify continuous variables; transmission values are reported band averages.}
    \label{tab:metamaterial-results}
    \begin{tabularx}{\linewidth}{@{}>{\RaggedRight\arraybackslash}m{0.34\linewidth}>{\RaggedRight\arraybackslash}m{0.25\linewidth}>{\RaggedRight\arraybackslash}m{0.20\linewidth}>{\columncolor{methodlabelfill}\Centering\arraybackslash}X@{}}
        \toprule
        Quantity / variable & Search range & Unit / context & Best value \\
        \midrule
        Optimization reward $R_{\mathrm{meta}}$ & -- & Dimensionless & $0.822$ \\
        Unit-cell period $p$ & $[5.0,9.0]$ & mm & $6.5$ \\
        Jerusalem-cross arm length $a_{\mathrm{jc}}$ & $[3.0,5.5]$ & mm & $3.95$ \\
        T-bar end-load length $b_{\mathrm{jc}}$ & $[0.4,1.8]$ & mm & $0.95$ \\
        Jerusalem-cross linewidth $w_{\mathrm{jc}}$ & $[0.2,0.5]$ & mm & $0.30$ \\
        Drude wire-grating pitch $p_w$ & $[1.0,3.0]$ & mm & $2.15$ \\
        Drude wire width $w_w$ & $[0.2,0.5]$ & mm & $0.255$ \\
        I-shaped inclusion neck length along $x$, $l_{\mathrm{neck}}$ & $[0.8,3.5]$ & mm & $0.81$ \\
        I-shaped inclusion pad length along $y$, $b_{\mathrm{pad}}$ & $[0.5,2.5]$ & mm & $1.60$ \\
        Total dielectric stack thickness $h_{\mathrm{total}}$ & $[3.0,6.0]$ & mm & $3.60$ \\
        Middle-layer thickness fraction $f_{\mathrm{mid}}$ & $[0.15,0.60]$ & Dimensionless & $0.435$ \\
        Average stopband transmission & -- & \%; 8--12 GHz & $6.2$ \\
        Average passband transmission & -- & \%; 14.5--15.5 GHz & $94.5$ \\
        \bottomrule
    \end{tabularx}
\end{table}

The expert-designed simple fishbone configuration had corresponding reported stopband and passband averages of 9.5\% and 87.4\%, compared with 6.2\% and 94.5\% for the \projectname{}-assisted design. The transmission spectra in Figure~\ref{fig:metamaterials}g show stopband suppression and high passband transmission across the five evaluated incidence angles. This case connects topology design with feedback-guided revision of the optimization space: the initial optimization exposed a restrictive thickness constraint, and the revised formulation enabled further exploration toward the two spectral objectives.

\FloatBarrier
\subsubsection{Alloy Design: Composition and Heat Treatment for Cast Magnesium}
\label{sec:alloy-design}

\textcolor{taiyiblue}{\textbf{Problem statement.}} The alloy collaboration seeks a high-tensile-strength magnesium alloy suitable for sand casting. The problem couples composition with solution treatment, quenching, and ageing: a promising nominal composition is useful only if a practical processing route can realize the desired mechanical properties. Cost also enters candidate selection, making the search a joint problem of performance, processing, and material economy.

\begin{figure}[p]
    \centering
\includegraphics[width=\linewidth,height=0.79\textheight,keepaspectratio]{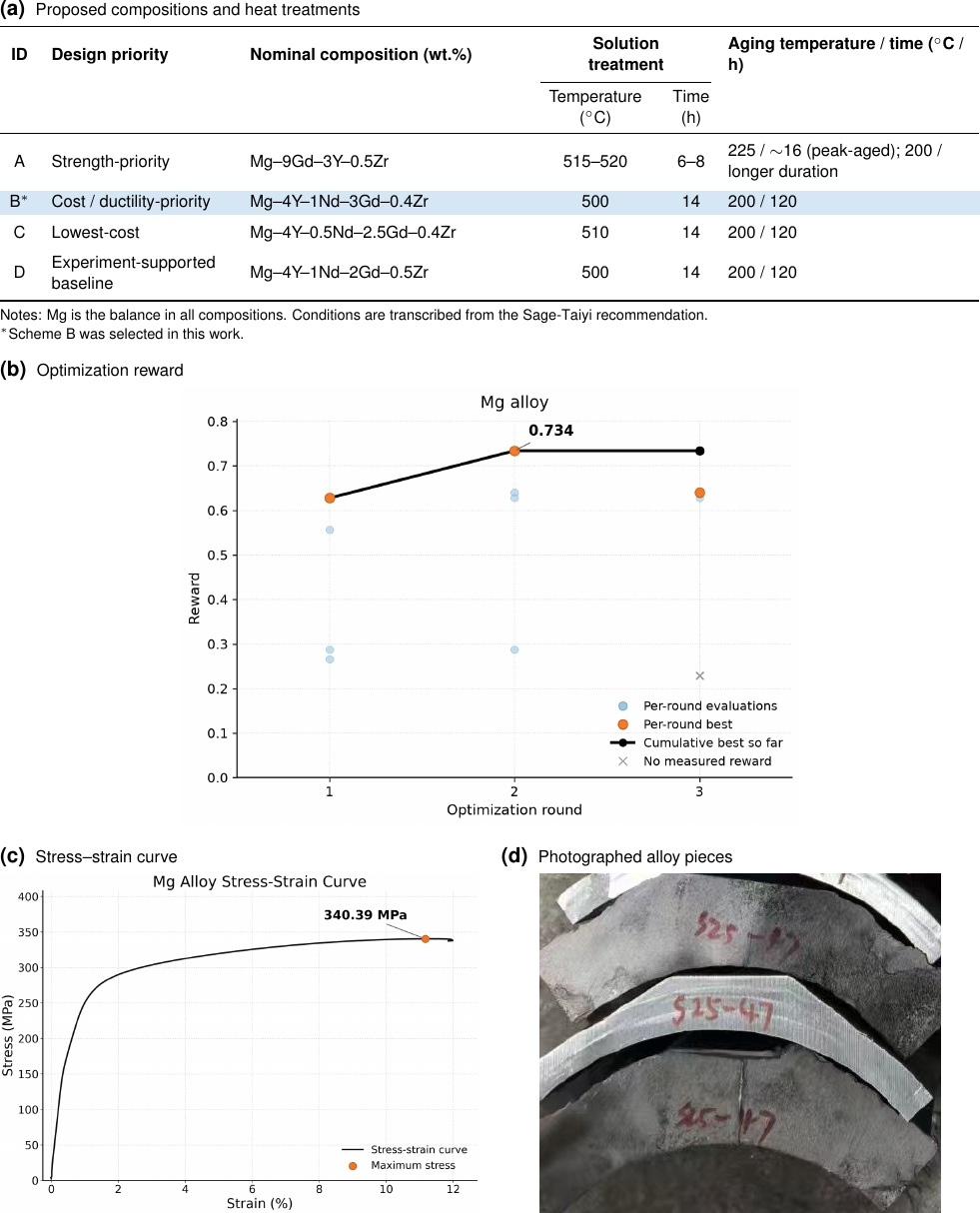}
    \caption{\textbf{Magnesium-alloy proposals, material, and evaluation.}
    \textbf{(a)} Four proposed compositions (wt.\%, Mg balance), design priorities, and heat-treatment schedules; the blue row and asterisk identify selected design B. The table records proposals.
    \textbf{(b)} Three-round reward history for the frozen composite objective in Eq.~\ref{eq:alloy-reward}; the best reward, 0.734, occurs in round two. The grey cross marks ``No measured reward'', not a measured value at its plotted height.
    \textbf{(c)} Supplied stress--strain curve from a representative test, with an annotated maximum of 340.39 MPa.
    \textbf{(d)} Photograph of the supplied alloy pieces. The collaboration reports 342 MPa as the mean tensile strength across repeated experiments; panel (c) shows one representative test at 340.39 MPa.}
    \label{fig:alloy-design}
\end{figure}

\textcolor{taiyiblue}{\textbf{Implementation.}} \projectname{} proposed four composition--processing candidates with different design priorities: strength-priority A, cost/ductility-priority B, lowest-cost C, and experiment-supported baseline D (Figure~\ref{fig:alloy-design}a). Experts selected design B, Mg--4Y--1Nd--3Gd--0.4Zr, as the starting point for experimentation. Its proposed heat treatment comprised solution treatment at 500\,$^{\circ}$C for 14 h followed by ageing at 200\,$^{\circ}$C for 120 h. Compositions are expressed in mass percentages (wt.\%), with Mg as the balance.

\begin{samepage}
Following the research harness described in Section~\ref{sec:real-world-feedback-method}, Taiyi-Optimizer jointly explored alloy composition and heat-treatment conditions, with physical synthesis and mechanical testing providing feedback for subsequent experimental choices. Cost and ductility informed the initial expert selection; the subsequent optimization maximized a fixed composite objective combining yield strength (YS), ultimate tensile strength (UTS), and elongation (EL):
\begin{equation}
    R_{\mathrm{alloy}}=0.35\,n(\mathrm{YS};110,210)
    +0.40\,n(\mathrm{UTS};210,345)
    +0.25\,n(\mathrm{EL};3,19),
\label{eq:alloy-reward}
\end{equation}
where
\begin{equation}
    n(x;l,h)=\operatorname{clip}\!\left(\frac{x-l}{h-l},0,1\right).
\label{eq:alloy-normalization}
\end{equation}
The function $n(x;l,h)$ normalizes a measured property $x$ using fixed lower and upper reference bounds $l$ and $h$, with $\operatorname{clip}$ limiting the result to $[0,1]$. YS and UTS are measured in megapascals (MPa), and EL in percent. The objective is maximized, with $R_{\mathrm{alloy}}\in[0,1]$. The search varied the nominal yttrium ($Y$), neodymium ($Nd$), gadolinium ($Gd$), and zirconium ($Zr$) contents together with solution-treatment and ageing temperatures and durations. Table~\ref{tab:alloy-results} lists the discrete search choices, units, and final configuration.

\end{samepage}

\textcolor{taiyiblue}{\textbf{Results.}} The best saved nominal configuration, its heat-treatment sequence, and the experimental outcomes are summarized in Table~\ref{tab:alloy-results}:
\begin{table}[H]
    \centering
    \fontsize{10}{12}\selectfont
    \renewcommand{\arraystretch}{1.02}
    \setlength{\tabcolsep}{4pt}
    \setlength{\parskip}{0pt}
    \renewcommand{\tabularxcolumn}[1]{m{#1}}
    \caption{Magnesium-alloy search choices, best recorded composition and processing conditions, and experimental outcomes. Discrete choices are shown as sets; nominal and measured compositions, and repeated-test and single-test results, are distinguished.}
    \label{tab:alloy-results}
    \begin{tabularx}{\linewidth}{@{}>{\RaggedRight\arraybackslash}m{0.34\linewidth}>{\RaggedRight\arraybackslash}m{0.25\linewidth}>{\RaggedRight\arraybackslash}m{0.20\linewidth}>{\columncolor{methodlabelfill}\Centering\arraybackslash}X@{}}
        \toprule
        Quantity / variable & Search range & Unit / context & Best value \\
        \midrule
        Optimization reward $R_{\mathrm{alloy}}$ & -- & Dimensionless & $0.734$ \\
        Nominal yttrium content $Y$ & $\{1,\allowbreak 2,\allowbreak 3,\allowbreak 4,\allowbreak 5\}$ & wt.\% & $4$ \\
        Nominal neodymium content $Nd$ & $\{1,\allowbreak 2,\allowbreak 3,\allowbreak 4,\allowbreak 5\}$ & wt.\% & $1$ \\
        Nominal gadolinium content $Gd$ & $\{1,\allowbreak 2,\allowbreak 3,\allowbreak 4,\allowbreak 5\}$ & wt.\% & $2$ \\
        Nominal zirconium content $Zr$ & $\{0.1,\allowbreak 0.2,\allowbreak 0.3,\allowbreak 0.4,\allowbreak 0.5\}$ & wt.\%; Mg balance & $0.5$ \\
        Solution-treatment temperature & $\{490,\allowbreak 495,\allowbreak 500,\allowbreak 505,\allowbreak 510,\allowbreak 515,\allowbreak 520,\allowbreak 525\}$ & $^{\circ}\mathrm{C}$ & $500$ \\
        Solution-treatment duration & $\{8,\allowbreak 10,\allowbreak 12,\allowbreak 14\}$ & h & $14$ \\
        Subsequent ageing temperature & $\{0,\allowbreak 200\}$ & $^{\circ}\mathrm{C}$ & $200$ \\
        Ageing duration & $\{0,\allowbreak 24,\allowbreak 48,\allowbreak 78,\allowbreak 102,\allowbreak 120,\allowbreak 144,\allowbreak 168,\allowbreak 192,\allowbreak 216,\allowbreak 312,\allowbreak 336\}$ & h & $120$ \\
        Measured composition & -- & wt.\%; Mg balance & \makecell[c]{Mg--4.01Y\\--0.98Nd\\--1.95Gd\\--0.5Zr} \\
        Mean tensile strength & -- & MPa; repeated tests & $342$ \\
        Representative maximum stress & -- & MPa; single test & $340.39$ \\
        \bottomrule
    \end{tabularx}
\end{table}
The final nominal composition was Mg--4Y--1Nd--2Gd--0.5Zr, refined from the selected design B, with solution treatment at 500\,$^{\circ}$C for 14 h and ageing at 200\,$^{\circ}$C for 120 h. Physical synthesis and testing yielded a mean tensile strength of 342 MPa across repeated experiments; the representative stress--strain curve reaches 340.39 MPa (Figure~\ref{fig:alloy-design}c). The composite reward reached its best value of 0.734 in the second optimization round (Figure~\ref{fig:alloy-design}b). This case connects joint composition--processing design and expert selection to a physically realized alloy, with experimental feedback supporting further refinement.

\FloatBarrier
\subsubsection{Compressor Engineering: Centrifugal Compressor Optimization}
\label{sec:compressor}

\textcolor{taiyiblue}{\textbf{Problem statement.}} The compressor-engineering collaboration targets an efficiency improvement for an existing centrifugal compressor design. The relevant choices couple impeller aerodynamics, diffuser matching, and mechanical margins. A design that improves one objective can make another constraint more restrictive, so the initial ideation stage explored different tradeoffs before focusing on geometric optimization.

\begin{figure}[p]
    \centering
    \includegraphics[width=\linewidth,height=0.78\textheight,keepaspectratio]{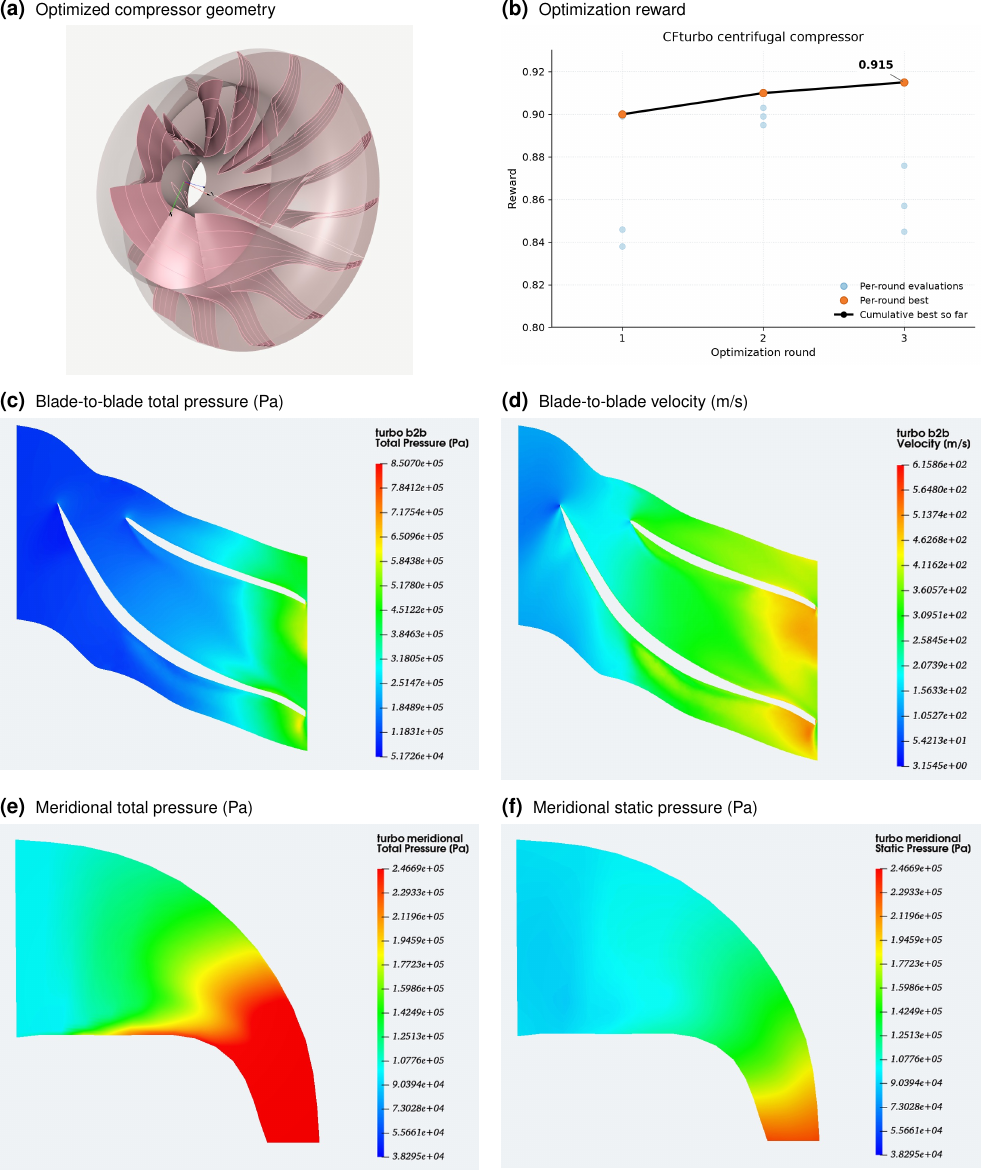}
    \caption{\textbf{Centrifugal-compressor geometry, optimization, and CFD fields.}
    \textbf{(a)} Rendering of the optimized compressor geometry.
    \textbf{(b)} Three-round reward history, reaching an annotated best of 0.915.
    \textbf{(c,d)} Blade-to-blade total pressure (Pa) and velocity (m/s).
    \textbf{(e,f)} Meridional total and static pressure (Pa).
    The revised source panels retain the complete colour scales and units. These panels contain a geometry rendering and simulation/optimization outputs; the reward is the impeller efficiency, with a final value of 0.915 (91.5\%) at PR $=4.0$.}
    \label{fig:compressor}
\end{figure}

\textcolor{taiyiblue}{\textbf{Implementation.}} \projectname{} proposed five routes addressing different aerodynamic and mechanical tradeoffs. The first combined a boreless Ti--6Al--4V backswept splitter impeller with a hybrid diffuser, targeting an efficiency greater than 0.76 while retaining aerodynamic and mechanical margins. The second combined a high-solidity impeller with $Z=18$ blades, a larger vaneless diffuser gap, and a boreless architecture to prioritize slip and structural margin. The third combined stronger backsweep and high solidity to reduce impeller-exit absolute Mach number and improve matching to a vaned diffuser. The fourth used inlet prewhirl, a wide-range vaneless or long-gap diffuser, and collector-gap injection to broaden the stable operating range, with tradeoffs in peak efficiency and mechanical margins. The fifth reduced the impeller outlet tip speed, $U_2$, to reduce centrifugal loading and fatigue risk, while accounting for diffuser matching and surge margin. Experts selected the first route for further development.

Initial design considerations covered outlet diameter and width, inlet shroud diameter, blade angles, and main-blade count. Following the workflow in Section~\ref{sec:real-world-feedback-method}, Taiyi-Optimizer searched the permitted geometric combinations, with computational fluid dynamics (CFD) evaluations informing subsequent parameter refinement. The final CFturbo search focused on four variables: outlet blade angle $\beta_{2b}$, inlet blade angle $\beta_{1b}$, main-blade count $Z_m$, and inlet shroud diameter $D_{1s}$. Blade angles follow the CFturbo convention and are expressed in degrees; $D_{1s}$ is expressed in millimetres. Table~\ref{tab:compressor-results} gives their search ranges and optimized values. Rotational speed and design mass flow were retained as design-point context. The resulting refinement reduced blade count and rebalanced the inlet and outlet angles.

\textcolor{taiyiblue}{\textbf{Results.}} The best saved configuration and reported CFD performance are summarized in Table~\ref{tab:compressor-results}:
\begin{table}[H]
    \centering
    \fontsize{10}{12}\selectfont
    \renewcommand{\arraystretch}{1.02}
    \setlength{\tabcolsep}{4pt}
    \setlength{\parskip}{0pt}
    \renewcommand{\tabularxcolumn}[1]{m{#1}}
    \caption{Compressor search ranges, best recorded configuration, and CFD performance. Blade count is discrete. The reward is impeller efficiency expressed as a fraction.}
    \label{tab:compressor-results}
    \begin{tabularx}{\linewidth}{@{}>{\RaggedRight\arraybackslash}m{0.34\linewidth}>{\RaggedRight\arraybackslash}m{0.25\linewidth}>{\RaggedRight\arraybackslash}m{0.20\linewidth}>{\columncolor{methodlabelfill}\Centering\arraybackslash}X@{}}
        \toprule
        Quantity / variable & Search range & Unit / context & Best value \\
        \midrule
        Optimization reward & -- & Efficiency fraction & $0.915$ \\
        Outlet blade angle $\beta_{2b}$ & $[115.6,135.6]$ & $^{\circ}$ & $117.0$ \\
        Inlet blade angle $\beta_{1b}$ & $[145.9,161.9]$ & $^{\circ}$ & $153.0$ \\
        Main-blade count $Z_m$ & $\{12,\allowbreak 13,\allowbreak 14,\allowbreak 15,\allowbreak 16\}$ & Blades & $12$ \\
        Inlet shroud diameter $D_{1s}$ & $[118,132]$ & mm & $129.0$ \\
        CFD pressure ratio $\mathrm{PR}$ & -- & Dimensionless & $4.0$ \\
        CFD impeller efficiency & -- & \%; CFD result & $91.5$ \\
        \bottomrule
    \end{tabularx}
\end{table}
Across three optimization rounds, the cumulative best CFD impeller efficiency increased from approximately 90.0\% to 91.0\% and then 91.5\%, corresponding to a final reward of 0.915 (Figure~\ref{fig:compressor}b). The final design achieved a pressure ratio $\mathrm{PR}=4.0$, defined as the ratio of outlet to inlet pressure. Figure~\ref{fig:compressor}a shows the optimized geometry, while panels c--f show blade-to-blade total pressure and velocity, followed by meridional total and static pressure. This case develops system-level design alternatives into a selected compressor geometry, with CFD feedback guiding parameter refinement.

\noindent\begin{minipage}{\linewidth}
\textbf{Cross-case interpretation.} Across the five collaborations, \projectname{} develops scientific ideas and executable search formulations, while the research harness connects them to expert selection, optimization, and experimental or simulation-based evaluation. The resulting feedback guides subsequent research decisions, linking scientific exploration to practical advances in reactions, materials, and engineering designs.

\end{minipage}

\addtocontents{toc}{\protect\setcounter{tocdepth}{2}}
\FloatBarrier

%% file: section/related_work.tex
\section{Related Work}
\label{sec:related-work}

\paragraph{AI Scientists.}
AI research agents connect literature review, hypothesis generation, and experimentation in workflows that span computational research~\cite{lu2024aiscientist,yamada2025aiscientistv2,tang2025airesearcher,feng2026internagent} and laboratory discovery~\cite{ghareeb2026robin,yang2026qiushi,swanson2025virtual,liu2026large,li2026large}. These systems refine hypotheses or intermediate results through critique~\cite{gottweis2026coscientist,yang2026aris}, carry findings forward through persistent memory and shared representations~\cite{lyu2026evoscientist,xiong2026evosci,mitchener2025kosmos,weng2026deepscientist,jiang2026deep}, and use execution results or surrogate models fitted to experimental observations to guide further search~\cite{novikov2025alphaevolve,yu2026largediscovery,si2026executiongrounded}.
Work focused on scientific ideation integrates evidence from prior research, retrieves and recombines concepts, and checks proposals against existing work~\cite{unipat2026uniscientist,wang2024scimon,baek2025researchagent,ghafarollahi2025sciagents,zhao2026researchstudioidea}. Search methods further develop promising ideas by exploring the literature, tracking evolving idea lineages, or progressively specifying experimental details~\cite{wang2026flowpie,gumma2026ideagent,yang2025moosechem2}. Evaluations examine complementary aspects of this process, including the breadth of exploration~\cite{tang2026narrowscientific}, how experts assess ideas before and after implementation~\cite{si2025novelideas,si2026ideationexecution}, and how new proposals build on prior research~\cite{zhou2026ideagenomes}.
\projectname{} focuses on how scientific judgments formed along one branch can inform decisions elsewhere in the search. Its revisable Cognitive State retains these judgments together with their supporting evidence and the conditions under which they apply, allowing the agent to revisit them when choosing research targets and methods.

\paragraph{Cognitive Science-Inspired Agent Design.}
Classical research describes problem solving as search over representations and operators~\cite{newellsimon1972human}; metacognition concerns how this process is monitored and controlled~\cite{nelsonnarens1990metamemory,cox2005metacognition}. CoALA organizes memory, internal and external actions, and decision cycles into a cognitive architecture for language agents~\cite{sumers2024coala}. Generative Agents uses experience retrieval and reflection to support planning~\cite{park2023generative}. Reflexion retains task feedback as verbal reflections to improve subsequent attempts~\cite{shinn2023reflexion}.
Recent work also studies which reasoning methods agents use and how they select them. \emph{Cognitive Foundations for Reasoning and Their Manifestation in LLMs} analyzes reasoning in terms of cognitive elements, including metacognitive controls, knowledge representations, and transformation operations, and finds that models underuse some elements associated with successful reasoning~\cite{kargupta2025cognitive}. Chain of Mindset uses a meta-agent to select spatial, convergent, divergent, or algorithmic reasoning modes according to the evolving reasoning state~\cite{jiang2026mindset}.
Together with cognitive-science research on creative thinking, metacognition, and scientific reasoning~\cite{mumfordmcintosh2017processes,benedek2023memory,gentner1983structuremapping,gickholyoak1983schema,flavell1979metacognition,nelsonnarens1990metamemory,ackermanthompson2017metareasoning,liedergriffiths2017strategy,fleming2024metacognition,klahrdunbar1988dual,klahrsimon1999discovery,kulkarni1988processes}, these approaches inform the design of \projectname{}'s Cognitive Repertoire and state-dependent method selection. \projectname{} maintains methodological heuristics and Cognitive Operators in a Cognitive Repertoire and instantiates them for research domains. Monitoring updates the Cognitive State from search outcomes; control uses that state and the repertoire to select targets and methods, revise the Target Tree, and guide idea evolution in the Idea Graph.

%% file: section/conclusion.tex
\section{Conclusion}
\label{sec:conclusion}

We presented \projectname{}, a metacognition-inspired architecture that uses accumulated scientific understanding to regulate scientific search. Through monitoring and control, a revisable Cognitive State and a reusable Cognitive Repertoire guide target selection, idea development, and validation. On Taiyi-Bench, \projectname{} leads all eight metrics, achieving an Overall score of 54.0 compared with 47.9 for the strongest baseline. Five expert-guided collaborations further illustrate how its ideas and search formulations connect to experimental and computational validation across chemistry, materials science, and engineering.

\enlargethispage{\baselineskip}
Future work will refine the Cognitive Repertoire through accumulated research experience and examine its reuse across domains. Building on this repertoire, \projectname{} will scale multi-agent collaboration to explore a broader range of high-value solutions within limited time and iteration budgets, aiming to improve the efficiency and quality of scientific discovery.

%% file: section/appendix.tex
\section{Contributions and Acknowledgments}
\label{app:team-members}

\subsection*{Lead authors}
Weida Wang, Benteng Chen, Yigeng Jiang, Shuchen Pu, Weiping Shen, Zichao Liu, Yulin Su, Zitong Wang, Siyu Zhou, Zifu Wang

\subsection*{Contributors}
 Yizhi Li, Xin Zong, Yuning Yang, Jixiang Bi, Yuan Wang, Chuyi Peng, Shiyao Geng, Qi Zhang, Kuan Lu, Mengzhu Li, Weiwang Xu, Peishan Xiang, Chengjun Liu, Chengrui Wei, Fangchen Yu, Shengji Tang, Yichen Jiang, Jingqi Ye, Cheng Yang, Tao Hu, Jianbiao Mei, Qinglong Cao, Kun Li, Yixin Chen, Shengyuan Xu, Juncheng Xiao, Yiheng Wang, Ruizhe Chen, Shiyang Feng, Runmin Ma, Xiangchao Yan, Zongsheng Cao, Haomeng Hu, Yu Ke, Taihui Wu, Meng Dai, Xu Guo, Haoyang Peng, Yinglong Ni, Pingsen Shi

\subsection*{Scientific Directors}
Xiaosong Wang, Shuyue Hu, Haijun Li, Sherry Dongqi Bao, Yang Yang, Chao Yang, Jie Hou, Tianfan Fu, Yi Gong, Jin Zeng, Mao Su, Hongxia Hao, Fanyang Mo, Fenghua Ling, Tianshu Yu, Yunqi Cai, Baihong Chi, Yuelu Chen, Jingwei Wang, Lei Wang, Chuncheng Bu, Shide Lv, Guangyu Wang, Yuqiang Li, Chunfeng Song, Lei Bai, Hui Wang, Xujun Fu

\subsection*{Advisors}
Li Cao, Dahua Lin, Xing Xie, Bowen Zhou

\subsection*{Core Directors}
Wenlong Zhang, Pinlong Cai, Peng Ye, Aiwen Lei, Jie Li, Shengchun Wang

\subsection*{Project Leader}
Bo Zhang, Shufei Zhang